\documentclass[conference]{IEEEtran}
\IEEEoverridecommandlockouts

\usepackage[noend]{algpseudocode}

\makeatletter
\def\BState{\State\hskip-\ALG@thistlm}
\makeatother

\newtheorem{theorem}{Theorem}

\usepackage{cite}
\makeatletter
\newcommand{\removelatexerror}{\let\@latex@error\@gobble}
\makeatother
\usepackage{amsmath,amssymb,amsfonts}
\usepackage{graphicx}
\usepackage{textcomp}
\usepackage{xcolor}
\usepackage{hyperref}
\DeclareMathOperator*{\argmin}{argmin}

\def\BibTeX{{\rm B\kern-.05em{\sc i\kern-.025em b}\kern-.08em
    T\kern-.1667em\lower.7ex\hbox{E}\kern-.125emX}}
\begin{document}

\title{Learning to Generate Synthetic Training Data using Gradient Matching and Implicit Differentiation
\\
}

\author{\IEEEauthorblockN{1\textsuperscript{st} Dmitry Medvedev}
\IEEEauthorblockA{\textit{Lomonosov MSU}\\
Moscow, Russia \\
dm.medvedev97@gmail.com}
\and
\IEEEauthorblockN{2\textsuperscript{nd} Alexander D'yakonov}
\IEEEauthorblockA{\textit{Lomonosov MSU} \\
Moscow, Russia
\\  djakonov@mail.ru
}
}

\maketitle

\begin{abstract}
Using huge training datasets can be costly and inconvenient. This article explores various data distillation techniques that can reduce the amount of data required to successfully train deep networks. Inspired by recent ideas, we suggest new data distillation techniques based on generative teaching networks, gradient matching, and the Implicit Function Theorem. Experiments with the MNIST image classification problem show that the new methods are computationally more efficient than previous ones and allow to increase the performance of models trained on distilled data.
\end{abstract}

\begin{IEEEkeywords}
data distillation, gradient matching, implicit differentiation, generative teaching network.
\end{IEEEkeywords}

\section{Introduction}
In machine learning, the purpose of data distillation~\cite{l1} is to compress the original dataset while maintaining the performance of the models trained on it. Generalizability is also needed: the ability of the dataset to train models of architectures that were not involved in the distillation process. Since training with less data is usually faster, distillation can be useful in practice. For example, it can be used to speed up a neural architecture search (NAS) task. Acceleration is achieved through the faster training of candidates.

In many recent works~\cite{l1},~\cite{l15},~\cite{l8},~\cite{l3},~\cite{l17}, distillation is formulated as an optimization problem with the objects of a new dataset as parameters for optimization. Therefore, to distill the dataset for an image classification task, pixels of images have to be optimized. First, all new objects are initialized with random noise, then these objects are used to train the student (randomly selected network). Then the student misclassification loss is calculated on real data. Finally, a gradient descent step is used to update the synthetic objects. Gradients can be calculated by backpropagating the error through the entire student's learning process. The step of this procedure can be very time-consuming and memory-intensive, so there is a need for an alternative. In~\cite{l14}, the authors use the implicit function theorem to solve the memory consumption problem. In~\cite{l15}, the data distillation problem has been reformulated to use gradient matching loss and speed up the optimization of synthetic objects and reduce memory usage.

There is an alternative to optimizing the pixels of synthetic data. In~\cite{l16}, the authors suggest to optimize parameters of the generator model (generative teaching network or GTN) to produce synthetic data from noise and labels. This creates a dataset that provides better performance for models trained with it. The disadvantage is that the authors used backpropagation through the learning process for optimization. Inspired by recent ideas in the field of data distillation, we propose replacing it with gradient matching or with implicit differentiation to make the procedure less computationally expensive. We have found that this allows not only to reduce memory costs but also to create more efficient and generalizable datasets. In addition, we investigate the use of augmentation in the distillation procedure and in models' learning on distilled data.

The rest of the paper is divided into 8 sections. We first overview the related work in section~\ref{rel_wor} and give a general formulation of the data distillation problem in~\ref{sec_gen}. We then analyse the first data distillation algorithm~\cite{l1} and discuss its problems in section~\ref{big_backprop}. A brief description of the algorithms for implicit differentiation~\cite{l14} and gradient matching~\cite{l15} can be found in sections~\ref{impl_diff} and~\ref{grad_match}.~\ref{gtn_section} presents the generative teaching network architecture that we use in our work. The~\ref{exps} section contains the results of experiments with the MNIST image classification benchmark. In~\ref{exp_tl} we compare the results of all the described distillation methods, limiting the distillation time to a constant. In~\ref{exp_gm} and~\ref{exp_ift} we show results of new distillation techniques when training a generator with gradient matching and implicit differentiation, respectively. In~\ref{exp_aug} we study the use of augmentation by distillation, and in~\ref{exp_gen} we check the generalization of the data obtained with the new methods. Finally, we present our findings in section~\ref{conc}. All code can be found in our GitHub page\footnote{\url{https://github.com/dm-medvedev/EfficientDistillation}}.

\section{Related Work}\label{rel_wor}
The general idea behind data distillation is to optimize hyperparameters (pixels of synthetic images are hyperparameters in an image classification problem) using gradients (also called hypergradients). The use of backpropagation~\cite{l6} to optimize hyperparameters has been suggested in~\cite{l9} and~\cite{l10}. Backpropagation through L-BFGS~\cite{l11} and SGD with momentum~\cite{l12} has been introduced in~\cite{l17}. Because of the great spatial complexity of this backpropagation, a more efficient one has been proposed in~\cite{l8}. There were also the results of data optimization experiments.

The successful distillation of the MNIST dataset~\cite{l2} was shown in~\cite{l1}. Leaving only 10 examples (one for each class), and thus reducing the dataset volume by 600 times, the LeNet model~\cite{l3} trained on compressed dataset showed an accuracy close to that of training on the original dataset. The authors also mentioned the distilled data generalization problem and suggested using a fixed distribution to initialize the network.

In~\cite{l3}, the authors show a way to distill both objects and their labels. Their experiments show that such distillation increases accuracy for multiple image classification benchmarks and allows distilled datasets to consist of fewer samples than number of classes. Despite this, recent works~\cite{l15},~\cite{l16} do not use label distillation, because joint optimization complicates the problem since labels depend on objects, and vice versa.

It is important to note that most of the works in data distillation were inspired by network distillation~\cite{l2}, that is the transfer of knowledge from an ensemble of well-trained models into a single compact one.

\section{General formulation}\label{sec_gen}
Let $\lambda$ be teacher parameters. These can be either GTN network's parameters, or synthetic objects' parameters (e.g. pixels of synthetic images).
To update $\lambda$, we must first train the student network $\theta$ on synthetic data, minimizing the task specific loss $\mathcal{L_S}$ (e.g. cross-entropy), and then get the loss on real data $\mathcal{L_T}$. To take care of generalizability, student's initialization goes from preset distribution $p(\theta_0)$. Afterall, the optimization problem for $\lambda$ can be formulated as follows:

\begin{align}\label{eq1}
&\lambda^* := \argmin\limits_{\lambda} \mathbb{E}_{\theta_0 \sim p(\theta_0)} \mathcal{L_T^*}\text{, where}\\
\nonumber
\mathcal{L_T^*} := &\mathcal{L_T}(\theta^*(\lambda)),\quad\quad
\theta^*(\lambda) := \argmin\limits_{\theta} \mathcal{L_S}(\lambda, \theta). 
\end{align}

To resolve the~\eqref{eq1} problem we can calculate gradient of $\mathcal{L_T}$ with respect to $\lambda$ to do the gradient descent step:

\begin{equation} \label{eq2}
  \frac{\partial \mathcal{L^*_T}}{\partial \lambda} = 
 \frac{\partial \mathcal{L_T}}{\partial \lambda} + \frac{\partial \mathcal{L_T}}{\partial \theta} \cdot \frac{\partial \theta^*}{\partial \lambda}
  =
  \frac{\partial \mathcal{L_T}}{\partial \theta} 
  \frac{\partial \theta^*}{\partial \lambda}
  .
\end{equation}

In this work we use cross-entropy loss as $\mathcal{L_T}$ and there is an explicit dependence only on $\theta$ and parameters of real data, so $\frac{\partial \mathcal{L_T}}{\partial \lambda} = 0$. Thus, the main part is the calculation of $\frac{\partial \theta^*}{\partial \lambda}$. Where the dependence of $\theta^*$ on $\lambda$ comes from student's training procedure. In our work, we use two methods of calculating~\eqref{eq2}: backpropagation through the student's learning process~\cite{l1} and implicit differentiation~\cite{l14}. Such a gradient can also be called hypergradient, since it is a gradient with respect to $\lambda$, which is a set of hyperparameters in the original student learning problem.

\section{Backpropagation through the student's learning process}\label{big_backprop}
This data distillation algorithm was suggested in~\cite{l1} and it is based on the assumption that the student's learning procedure is differentiable. This means that we can backpropogate gradient through it. We will denote it as \textbf{unroll}. Let $\theta_i$ be the student's parameters obtained at the i-th step of the training procedure, $\mathcal{B^T}$ be a batch of original data, and $\eta$ be the learning rate, then:

\begin{align} \label{eq:3}
\nonumber
&\theta_0 \sim p(\theta_0);\\
&\theta_{n+1} = \theta_{n} - \eta \nabla_{\theta} \mathcal{L_S}(\lambda, \theta_n); 
\quad k = 0, ..., N-1;\\
\nonumber
&\mathcal{L_T} = \textit{ClassificationLoss}(\mathcal{B^T}, \theta_N(\lambda)) \rightarrow \min\limits_{\lambda}.
\end{align}

\begin{figure}
\begin{algorithmic}[1]
\State \textbf{Input:} teacher's parameters $\lambda$, student's initialization distribution $p(\theta_0)$, number of distillation epochs $K$, number of student's learning steps $N$, real data $\mathcal{T}$, learning rate $\eta$.
\For{$k = 1,...,K$}:
\State $\mathcal{B^T} \sim \mathcal{T}, \quad \theta_0 \sim p(\theta_0)$ \Comment{sample batch and weights}
\State \textbf{Memory} $\leftarrow \theta_0$ \Comment{store initial weights}
\For{$n = 0,...N-1$}:
\State $g_n = \eta \frac{\partial\mathcal{L_S}(\lambda, \theta_n)}{\partial \theta_n}$
\State $\theta_{n+1} = \theta_n - g_n$
\State \textbf{Memory} $\leftarrow$ $g_n, \theta_{n+1}$ \Comment{store graph and weights.}
\EndFor
\State $\mathcal{L_T} = \textit{ClassificationLoss}(\mathcal{B^T}, \theta_N(\lambda))$ 
\State $\nabla_{\lambda} \mathcal{L_T}$ $\leftarrow \textbf{hypergrad}_{\text{unroll}}(\textbf{Memory}, \mathcal{L_T})$ \Comment{\figurename~\ref{alg2}}
\State \textbf{Update}($\lambda, \nabla_{\lambda} \mathcal{L_T}$) \Comment{update with any optimizer}
\EndFor
\State \textbf{Output:} $\lambda$
\end{algorithmic}
\caption{Backpropagation through the learning process.}\label{alg1}
\end{figure}

\begin{figure}
\begin{algorithmic}[1]
\State \textbf{Input:} loss on real data $\mathcal{L_T}$, computational graph and weights $\textbf{Memory}$.
\State $\theta_N \leftarrow \textbf{Memory}, \quad v = \frac{\partial\mathcal{L_T}}{\partial\theta_N}, \quad \nabla_{\lambda} \mathcal{L_T} = 0$ 
\For{$n = N-1,...0$}:
\State $g_n, \theta_n \leftarrow \textbf{Memory}$
\State $\nabla_{\lambda} \mathcal{L_T} \mathrel{{-}{=}}
\textbf{grad}\big(
  \textbf{func}=g_n, 
  \textbf{wrt}=\lambda, 
  \textbf{vec}=v\big)$
\State $v \mathrel{{-}{=}} 
\textbf{grad}\big(
  \textbf{func}=g_n, 
  \textbf{wrt}=\theta_n, 
  \textbf{vec}=v\big)$
\EndFor
\State \textbf{Output:} $\nabla_{\lambda} \mathcal{L_T}$ 
\end{algorithmic}
\caption{$\textbf{hypergrad}_{\text{unroll}}(\textbf{Memory}, \mathcal{L_T})$.}\label{alg2}
\end{figure}

The learning rate $\eta$ can be optimized in the same way as $\lambda$, but in~\cite{l15} 
and~\cite{l17} it was found that this leads to overfitting of the synthetic dataset to the architecture of student used in distillaion process. To write out the desired derivative $\frac{\partial \theta^*(\lambda)}{\partial \lambda}$, we can unroll the learning procedure (see full derivation in~\cite{l14}):

\begin{align} \label{eq:4}
\frac{\partial \theta^*(\lambda)}{\partial \lambda} &=
\sum\limits_{1 \leq j \leq N} \Bigg[ \prod\limits_{1 \leq k < j} \Bigg(
I - \eta\frac{\partial^2 \mathcal{L_S}(\lambda, \theta_{N-k})}{\partial \theta^2} \Bigg)
\Bigg] \cdot  \\
\nonumber
&\cdot
\frac{\partial^2 \mathcal{L_S}(\lambda, \theta_{N - j})}{\partial \theta \partial \lambda} \cdot (-1).
\end{align}

The resulting algorithm (see \figurename~\ref{alg1} and~\ref{alg2}) can be implemented using the Higher library~\cite{l18}. Note that Higher allows to backporopogate through many optimizers besides simple gradient descent. In our paper we use SGD with momentum~\cite{l12}. Note that \textbf{grad} in~\figurename~\ref{alg1} and~\ref{alg2} denotes Vector Jacobian product.

This distillation method is both time and space consuming. To perform a single step of updating $\lambda$ it is necessary to perform $N$ (see \figurename~\ref{alg1}) student optimization steps, while all intermediate results (copies of the student weights) must be stored in memory. Considering that usually a student's training can take many optimization steps, the efficiency problem become the main one. There is also a problem with the generalization of resulting syntetic dataset, which can be solved by sampling student's initialization and architecture. In our work we only randomly sample initializations.

Note that the procedure of student's training on the resulting synthetic dataset can be carried out in different ways. New data, parameterized with $\lambda$, can be used as single large batch or it can be split into several smaller ones. This split can be useful to reduce memory consumption per training step. Instead of randomly sample distilled objects, the authors of the original work propose to attach each of them to a specific batch. These batches can have a certain order in an epoch. In our paper, we use the same schemes, and in addition, we choose $K$ (see \figurename~\ref{alg1}) to stay within the particular time limit. Let $ic$ (input count) be the number of batches of the synthetic dataset, note that it must be divisor of $N$. In our experiments we try $ic=1$ and $ic=10$.

\section{Implicit Differentiation}\label{impl_diff}
This method suggested in~\cite{l14} is based on implicit function theorem:

\begin{theorem}[Cauchy, Implicit Function Theorem] 

Let $\frac{\partial \mathcal{L_S}}{\partial \theta}(\lambda, \theta): \Lambda \times \Theta \rightarrow \Theta$, be a continuously differentiable function. Fix a point $(\lambda^{'}, \theta^{'})$ with $\frac{\partial \mathcal{L_S}}{\partial \theta} (\lambda^{'}, \theta^{'}) = 0$. If the Jacobian matrix $\frac{\partial^2 \mathcal{L_S}}{\partial \theta^2}$ is invertible, then there exists an open set $\lambda: U \subseteq \Lambda$ containing $\lambda^{'}$ such that there exists a unique continuously differentiable function $\theta^*: U \rightarrow \Theta$, such that:
\begin{align}
\nonumber
\theta^*(\lambda^{'}) = \theta^{'} \quad \text{and}\quad 
\forall \lambda \in U, \quad \frac{\partial \mathcal{L_S}}{\partial \theta} (\lambda, \theta^*(\lambda)) = 0.
\end{align}
Moreover, the partial derivatives of $\theta^*$ in $U$ are given by the matrix product:
\begin{align} \label{eq6}
\frac{\partial \theta^*}{\partial \lambda} (\lambda) = - 
\Bigg[\frac{\partial^2 \mathcal{L_S}}{\partial \theta^2} (\lambda, \theta^{*}(\lambda))\Bigg]^{-1} \frac{\partial^2 \mathcal{L_S}}{\partial \theta \partial \lambda} (\lambda, \theta^{*}(\lambda)).
\end{align}
\end{theorem}

So, if there was an efficient way to invert the matrix, we would simply use~\eqref{eq6}, after the student $\theta$ has reached a local minimum, assuming $\frac{\partial \mathcal{L_S}}{\partial \theta} (\lambda, \theta^*(\lambda)) \approx 0$. But the inversion operation is time costly, so the authors used the approximation by the Neumann series:

\begin{align} \label{eq7}
\Bigg[\frac{\partial^2 \mathcal{L_S}}{\partial \theta^2} (\lambda, \theta^{*}(\lambda))\Bigg]^{-1} =&\\
= \lim\limits_{i \rightarrow \infty} \sum\limits_{j=0}^i &\Bigg[ I -
\nonumber
\frac{\partial^2 \mathcal{L_S}}{\partial \theta^2}(\lambda, \theta^{*}(\lambda))\Bigg]^j.
\end{align}

To approximate the desired derivative, we just need to take the first few elements of the~\eqref{eq7} series. To ensure the convergence of the series, the maximum absolute eigenvalue of the matrix must be less than one. Therefore, the authors used the additional hyperparameter $\alpha$:

\begin{align}
\Bigg[\frac{\partial^2 \mathcal{L_S}}{\partial \theta^2} (\lambda, \theta^{*}(\lambda))\Bigg]^{-1} \approx& \\
\nonumber
\approx \alpha \sum\limits_{j=0}^N &\Bigg[ I - \alpha \frac{\partial^2 \mathcal{L_S}}{\partial \theta^2}(\lambda, \theta^{*}(\lambda))\Bigg]^j.
\end{align}

\begin{figure}
\begin{algorithmic}[1]
\State \textbf{Input:} teacher's parameters $\lambda$, student's initialization distribution $p(\theta_0)$, number of distillation epochs $K$, number of student's learning steps $\zeta_\theta$, real data $\mathcal{T}$, learning rate $\eta$.

\For{$k = 1, ..., K$}
\State $\mathcal{B^T} \sim \mathcal{T}, \quad \theta \sim p(\theta_0)$
\For{$n = 1, ..., \zeta_\theta$}
\State $\theta \mathrel{{-}{=}} \eta \frac{\partial \mathcal{L_S}(\lambda, \theta)}{\partial\theta}$
\EndFor
\State $\mathcal{L_T} = \textit{ClassificationLoss}(\mathcal{B^T}, \theta)$
\State $\nabla_{\lambda} \mathcal{L_T} = \textbf{hypergrad}_{\text{IFT}}(\mathcal{L_T}, \mathcal{L_S}, \lambda, \theta)$ \Comment{see \figurename~\ref{algo4}}
\State \textbf{Update}($\lambda, \nabla_{\lambda} \mathcal{L_T}$) \Comment{update with any optimizer}
\EndFor
\Return $\lambda$
\end{algorithmic}
\caption{Distillation with implicit differentiation.}\label{algo3}
\end{figure}

\begin{figure}
\begin{algorithmic}[1]
\State \textbf{Input:} loss on real data $\mathcal{L_T}$, loss on synthetic data $\mathcal{L_S}$, teacher’s parameters $\lambda$, student’s parameters $\theta$.
\State $p = v = \frac{\partial \mathcal{L_T}}{\partial \theta}$
\For{$j = 1,...,N$} \Comment{$N$ --- number of elements in~\eqref{eq7}}
\State $v \mathrel{{-}{=}} \alpha \cdot \textbf{grad}\big(\textbf{func}=\frac{\partial \mathcal{L_S}}{\partial \theta}, \textbf{wrt}=\theta, \textbf{vec}=v\big)$
\State $p \mathrel{{+}{=}} v$
\EndFor
\State\Return $- \alpha \cdot \textbf{grad}\big(\textbf{func}=\frac{\partial \mathcal{L_S}}{\partial \theta}, \textbf{wrt}=\lambda, \textbf{vec}=p\big)$
\end{algorithmic}
\caption{$\textbf{hypergrad}_{\text{IFT}}(\mathcal{L_T}, \mathcal{L_S}, \lambda, \theta)$.}\label{algo4}
\end{figure}

The resulting algorithm (see \figurename~\ref{algo3} and~\ref{algo4}) has no problems with memory consumption since there is no need to store copies of the student $\theta$. And, despite the many approximations in calculations, the experimental results show that method has a competitive performance (see Table~\ref{tab4}).

Another interesting detail of this method is that there is no dependence on which optimizer is used to train the student, and on the order (curriculum) of batches of synthetic data. So, in our paper we only use single large batch of synthetic data. The original work~\cite{l14} lacks a detailed description of the experimental results, so it can be found in our paper (see section~\ref{exp_ift}). We used the open-source code\footnote{\url{https://github.com/AvivNavon/AuxiLearn}} as the basis for the implementing the method.

\section{Gradient Matching}\label{grad_match}
The gradient matching method (\textbf{GM}) was proposed in~\cite{l15}, and it solves a different problem than the general one (see section~\ref{sec_gen}). The main difference is that we want not only to train the student $\theta$ to achieve a good performance on real data but also to get such a solution as if it was trained on real data. To formulate this let $D(\theta_1, \theta_2)$ be the function of how close one student's parameters are to another. Let $\theta^\mathcal{S}$ and $\theta^\mathcal{T}$ be parameters obtained by training on distilled and real data, respectively. $\zeta_{\mathcal{S}}$ and $\zeta_{\mathcal{T}}$ are the number of steps to train the student on synthetic and real data. The optimization of student is done with $opt_\theta$ (it can be any known optimization algorithm), then:

\begin{align}
&\lambda^* = \argmin_{\lambda} \mathbb{E}_{\theta_0 \sim p_{\theta_0}} \Big[ \sum\limits_{n=1}^N D(\theta_n^{\mathcal{S}}, \theta_n^{\mathcal{T}})\Big],\quad
\text{where: }\\
\nonumber
\theta^{\mathcal{S}}_t &= \text{opt}_{\theta}(\mathcal{L_S}(\lambda, \theta^{\mathcal{S}}_{t-1}), \zeta_{\mathcal{S}}), \quad \quad
\theta^{\mathcal{T}}_t = \text{opt}_{\theta}(\mathcal{L_T}(\theta^{\mathcal{T}}_{t-1}), \zeta_{\mathcal{T}}).
\end{align}

Let $D(\theta^{\mathcal{S}}_{t-1}, \theta^{\mathcal{T}}_{t-1}) \approx 0$, such assumption is true if we are close to the problem solution. Note that $\theta_n = \theta_{n-1} - \nabla_{\theta} \mathcal{L_{S}}$, then:

\begin{align}
\lambda^* = \argmin_{\lambda} \mathbb{E}_{\theta_0 \sim P_{\theta_0}} \Big[ \sum\limits_{n=1}^{N-1} D\big(\nabla_{\theta} \mathcal{L_{S}}(\lambda, \theta_{n}), \\
\nonumber
\nabla_{\theta} \mathcal{L_{T}}(\theta_{n})\big)\Big].
\end{align}

The distance function $D$ is just the sum (in our paper for GTN experiments we used the mean) of the cosine distance functions for each student layer $\theta^l$. Let $A$ and $B$ be gradient tensors with respect to layer parameters. Let $i$ be the index of the output axis (e.g. for a convolutional layer this is the index of the output channel). Let $A_i$ and $B_i$ be flat gradient vectors corresponding to each output element indexed by $i$ then:

\begin{align}\label{form_d}
D(\nabla_{\theta}\mathcal{L_S}, \nabla_{\theta}\mathcal{L_T}) &= \sum\limits^L_{l=1} d(\nabla_{\theta^l}\mathcal{L_S}, \nabla_{\theta^l}\mathcal{L_T}), \quad \text{where}\\
\nonumber
d(A, B) &= \sum^{\text{dim(A)}}_{i=1} \Bigg( 1 - \frac{A_i \cdot B_i}{\|A_i\| \|B_i\|} \Bigg).
\end{align}

The most interesting detail here is that the authors suggest to update $\lambda$ after each step of student optimization, so now we don't need to wait until it reaches a local minimum, as it was before. The authors also propose not to store student copies and to minimize $D\big(\nabla_{\theta} \mathcal{L_{S}}(\lambda, \theta_{t-1}), \nabla_{\theta} \mathcal{L_{T}}(\theta_{t-1})\big)$ for each step separately. So there is no backpropagation through $opt_{\theta}$. Both of these proposals make the gradient matching method very computational effective.

\begin{figure}
\begin{algorithmic}[1]
\State\textbf{Input}: teacher's parameters $\lambda$ and synthetic objects $\mathcal{S}(\lambda)$, student's initialization distribution $p(\theta_0)$, number of distillation epochs $K$, number of student's learning steps $\zeta_\theta$, real data $\mathcal{T}$, learning rate $\eta_{\theta}$, number of inner loop steps $N$.

\For{$k = 0, ..., K-1$}
\State $\theta_0 \sim p_{\theta_{0}}$
\For{$n = 0, ..., N-1$}
\State $\mathcal{B^T} \sim \mathcal{T}, \quad \mathcal{B^S} \sim \mathcal{S}(\lambda)$
\State $\mathcal{L_{T}} = \textit{ClassificationLoss}(\mathcal{B^T}, \theta_n)$
\State $\mathcal{L_{S}} = \textit{ClassificationLoss}(\mathcal{B^S}, \theta_n)$
\State $\mathcal{L}(\lambda) = D(\nabla_{\theta} \mathcal{L_S}(\lambda, \theta_n), \nabla_{\theta} \mathcal{L_T}(\theta_n))$
\State \textbf{Update}$(\lambda, \nabla_\lambda \mathcal{L}(\lambda))$
\State $\theta_{n+1} \leftarrow opt_{\theta}(\mathcal{L_S}(\lambda, \theta_n), \zeta_{\theta}, \eta_{\theta})$
\EndFor
\EndFor
\State\textbf{Output}: $\lambda$
\end{algorithmic}
\caption{Gradient matching.}
\end{figure}

The peculiarity of this loss function is that the gradient of one synthetic object depends on other objects from the same batch, because of a normalization operation in the $d$ equation~\eqref{form_d}. It makes the optimization problem harder and can cause negative effects (see Table~\ref{tab2}). So authors decided to distill objects separately for each class.

Note that gradient matching is independent of the student training optimization algorithm. There is only one assumption that the direction should be based on the gradient. Another detail is that the curriculum (the order of the synthetic batches in the student's learning procedure) can be learned with this distillation method. We used open-source code\footnote{\url{https://github.com/VICO-UoE/DatasetCondensation}} as the implementation of this method.

\section{Generative Teaching Network}\label{gtn_section}

The idea first appeared in~\cite{l16}, where authors suggested to use the generator as the teacher $\lambda$. The input of the generator is a concatenation of noise and one hot encoded label (for conditional generation). In the original paper, the authors use backpropagation through the student's learning process to train the generator, which is inconvenient for practical use due to high memory consumption, so in our paper, we show that the same or even better results can be achieved more effectively by using gradient matching or implicit differentiation.

Experimental results in~\cite{l16} show that using a generator can help improve student performance. The best results were achieved with the learned curriculum. This was done by treating the generator input as teacher parameters and fixing their order. Thus, the generator produces only a finite number of synthetic objects and gives them for training the student as batches in a fixed order. This makes sense since the use of a generator can be seen as a more general case of usual distillation (when the parameters of objects are optimized).

If the generator input is synthetic images and the generation operation is the product of the images and generator parameters, which are the identity matrix, then there will be the usual data distillation. In our paper, we check if we can improve distillation performance using larger generators.

\begin{figure}[h!]
\begin{minipage}[h]{\linewidth}
\centering
\includegraphics[width=0.4\linewidth]{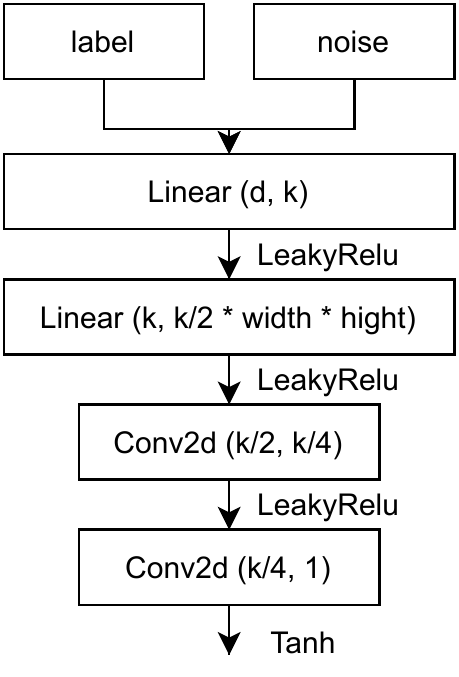}
\end{minipage}
\caption{Generator's architecture. $k$ --- hyperparameter to control network's size. $d=64$ --- generator's input.}\label{ArchRis}
\end{figure}

Note that the size in our experiments is controlled by the \textit{k} hyperparameter (see \figurename~\ref{ArchRis}). The generator consists of two linear layers and two convolutional layers. The output size of the first layer is \textit{k}. And $\lfloor k/2 \rfloor \times \text{(width)} \times \text{height of picture}$ is the output size of the second layer. $\lfloor k/4\rfloor$ is the number of output channels of the first convolution. 

Hereinafter, unless otherwise indicated, we use the following notation: \textbf{DD} (data distillation) --- distillation, when the parameters of the teacher $\lambda$ are pixels of synthetic images, and \textbf{GTN} --- for distillation using a generator. Note that the generator has two modes: \textbf{GTN-rnd} --- generator with random noise as input, (\textbf{GTN-lrn}) --- generator with learned input.

\section{Experiments}\label{exps}
\subsection{Distillation with time limit}\label{exp_tl}
The neural architecture search (NAS) is one of the most promising areas for distillation and it is important to note that the time spent on distillation should be added to the time spent on the NAS, this idea was also mentioned in review\footnote{\url{https://openreview.net/forum?id=HJg_ECEKDr}} of~\cite{l16}. So, in this section, we check the performance of all known distillation methods. We think that it is fair to distill the data by all methods for the same limited time. We have chosen a time limit of $\approx 15$ minutes, and it is based on common sense and NAS time spent in similar experiments~\cite{l15}. Note that this limit may not be accurate, as distillation takes an integer number of steps, and each step may take slightly different times.

To check the performance we use the following scheme. First we train teacher $\lambda$ with three restarts. The number of steps is determined by the time limit indicated above. Then, to get the final results we train five randomly initialized students $\theta$ for each of the three teachers. Each student's training takes 1000 optimization steps.

In our work we use the MNIST~\cite{l4} benchmark and make the same preparations as in~\cite{l16}. We extract part of the training data for validation (10 thousands of images) and use it to get the best teacher hyperparameters. We use $|\mathcal{B^T}|=256$ batch size of training data. For most of our experiments we use ConvNet~\cite{l19} as a student. As student's optimizer we use SGD with momentum with the same parameters as suggested in~\cite{l15}. We use the same teacher optimizers as in the original papers~\cite{l1},~\cite{l15},~\cite{l16}. The volume of synthetic data can be controlled by \textit{ipc} (images per class) parameter. For each table in this paper, the largest numbers in the column are shown in bold.

\begin{table}[h!]
\caption{Mean and standard deviation of test accuracy for different distillation algorithms.}
 \centering
 \begin{tabular}{l|c|c|c}
Method + Teacher & Accuracy & Params & GPU (MiB) \\\hline
GM + DD ($K=60,$& \boldmath{$94.9 \pm 0.1$} & $78.4$ K & $\approx 2390$\\
$\hspace{1.5cm}\zeta_\theta=50$)& & & \\\hline
unroll + DD ($ic=1$) & $88.4 \pm 0.3$ & $78.4$ K & $\approx 4432$ \\\hline
unroll + DD ($ic=10$) & $79.2 \pm 0.7$ & $784$ K & $\approx 4426$ \\\hline
unroll + GTN-lrn & $92.0 \pm 0.3$ & $1.646$ M & \boldmath{$\approx 4480$} \\
($ic=1$) &&&\\\hline
unroll + GTN-lrn & $91.6 \pm 0.5$ & \boldmath{$1.704$} M & \boldmath{$\approx 4480$} \\
($ic=10$) &&&\\\hline
unroll + GTN-rnd & $91.7 \pm 0.3$ & $1.640$ M & \boldmath{$\approx 4480$} \\\hline
 \end{tabular}
\label{tab1}
\end{table}

Table~\ref{tab1} shows mean and standard deviation of test accuracy, reached by students trained on distilled data. Note that there is only one difference from previous works, we use time limit for each distillation procedure, so there is degradation in performance. For this experiment, we use $K=1000, N=10$ as default hyperparameters values.

To check the memory consumption we use a special tool\footnote{\url{https://pytorch.org/docs/stable/cuda.html\#torch.cuda.max_memory_reserved}}, which can measure the GPU memory usage. Note that using of the \textbf{unroll} distillation procedure consumes the most memory. The second column shows the number of teacher parameters, and although \textbf{GTN} ($k=64$) is twice as large as \textbf{DD}, there is not much difference in memory usage.

\subsection{Training generator with gradient matching}\label{exp_gm}
\begin{figure}[h!]
\begin{minipage}[h]{0.48\linewidth}
\centering
\includegraphics[width=\linewidth]{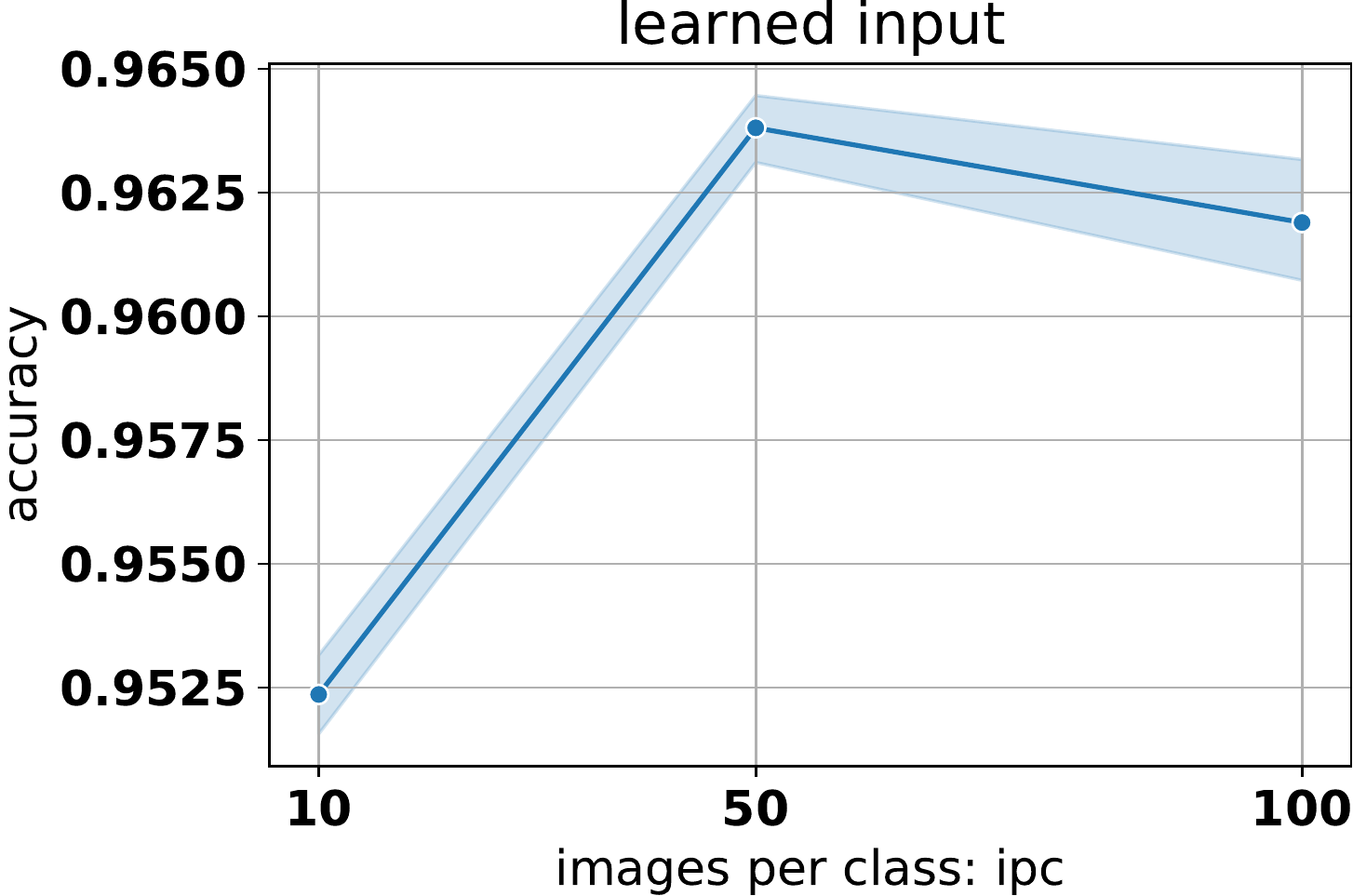}\\a)
\end{minipage}
\hfill
\begin{minipage}[h]{0.48\linewidth}
\centering
\includegraphics[width=\linewidth]{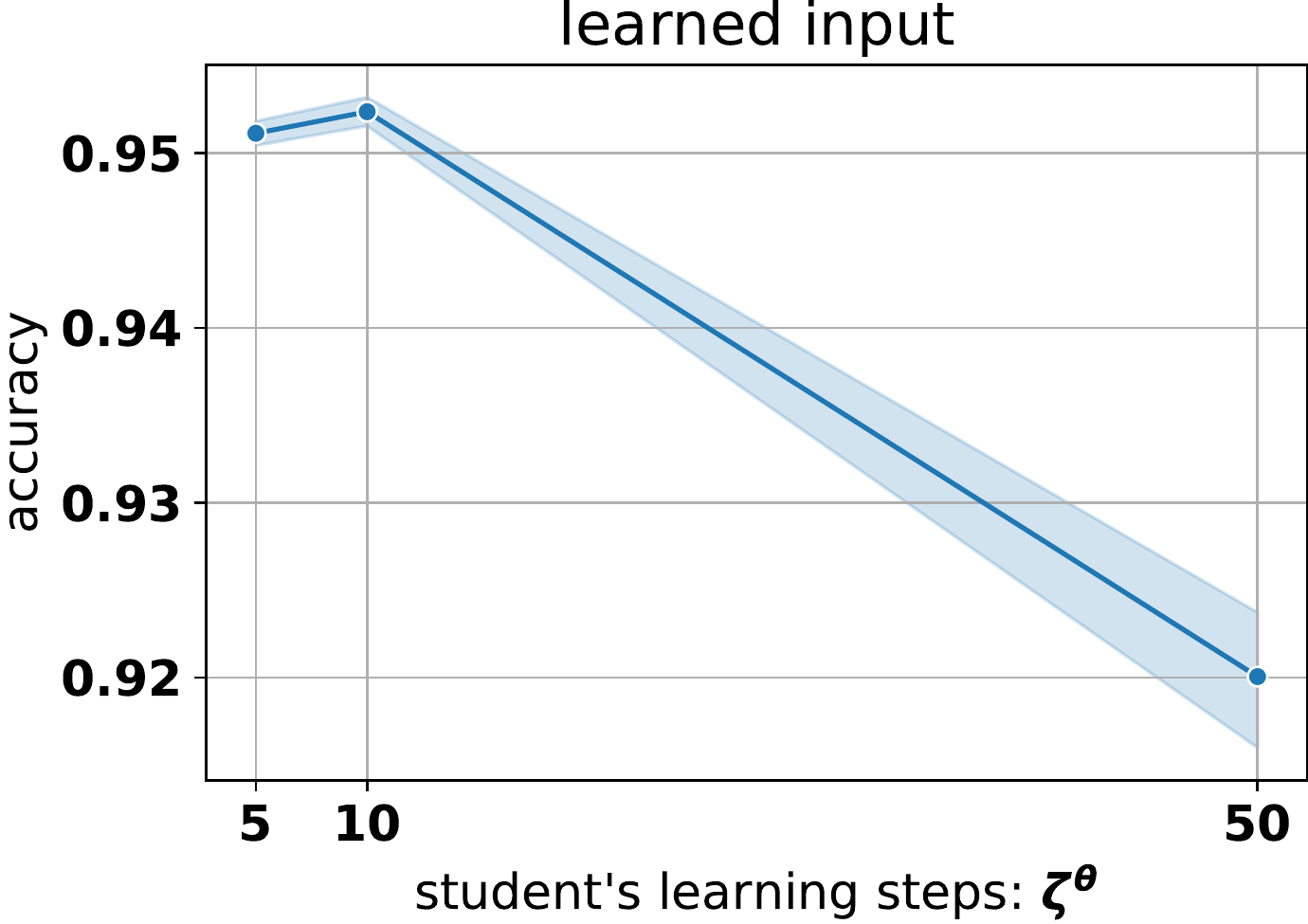}\\b)
\end{minipage}
\vfill
\begin{minipage}[h]{0.48\linewidth}
\centering
\includegraphics[width=\linewidth]{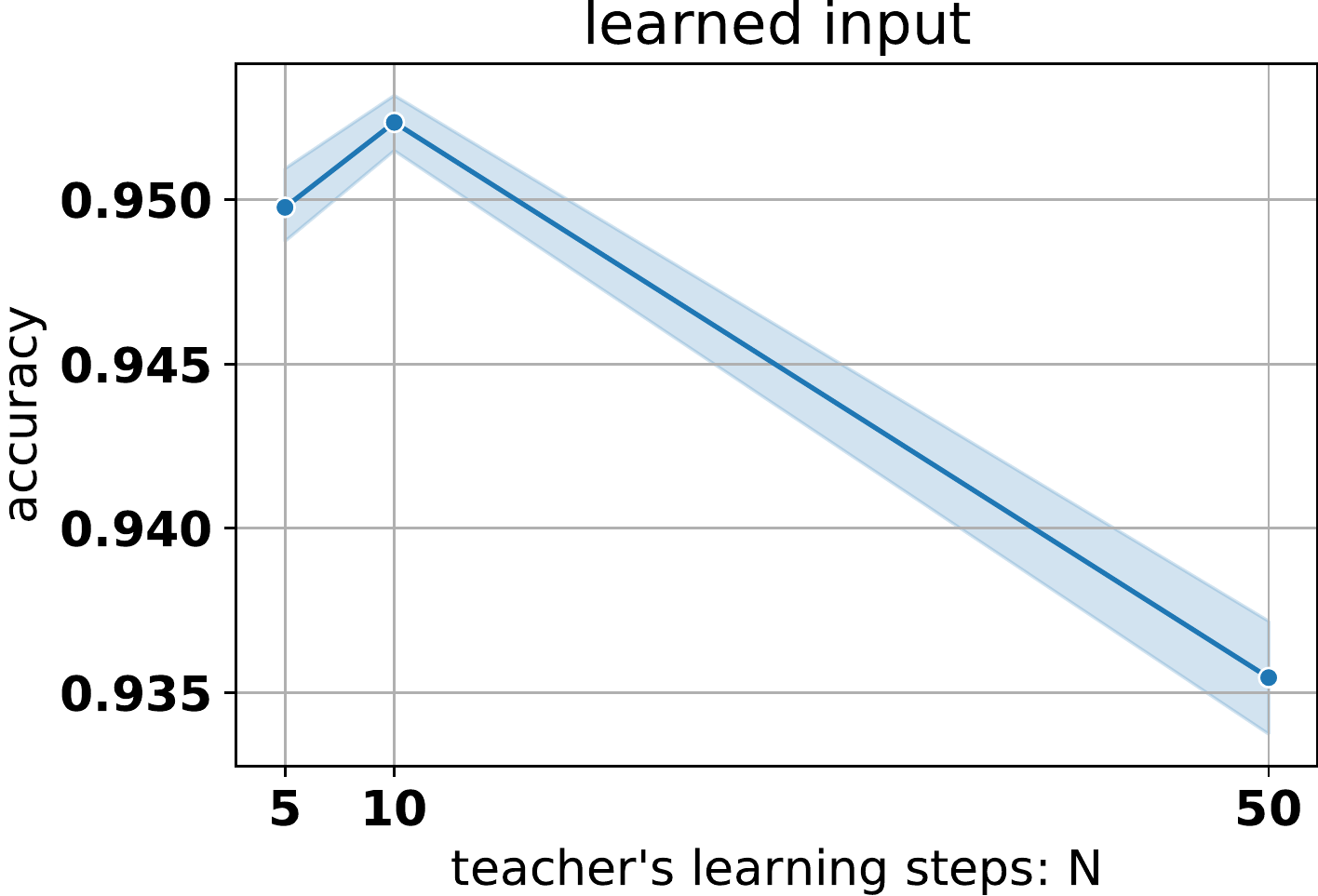}\\c)
\end{minipage}
\hfill
\begin{minipage}[h]{0.48\linewidth}
\centering
\includegraphics[width=\linewidth]{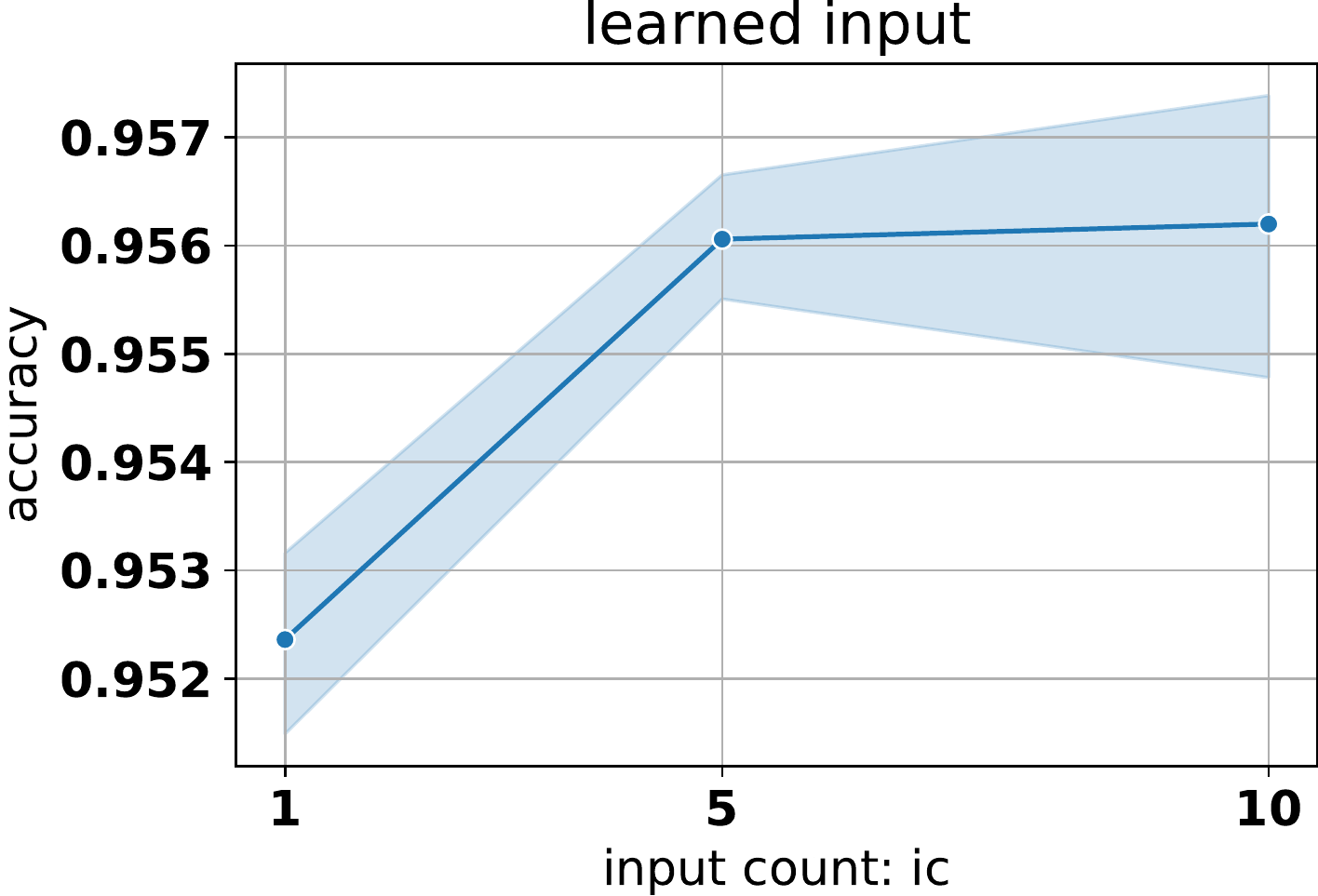}\\d)
\end{minipage}
\vfill
\begin{minipage}[h]{0.48\linewidth}
\centering
\includegraphics[width=\linewidth]{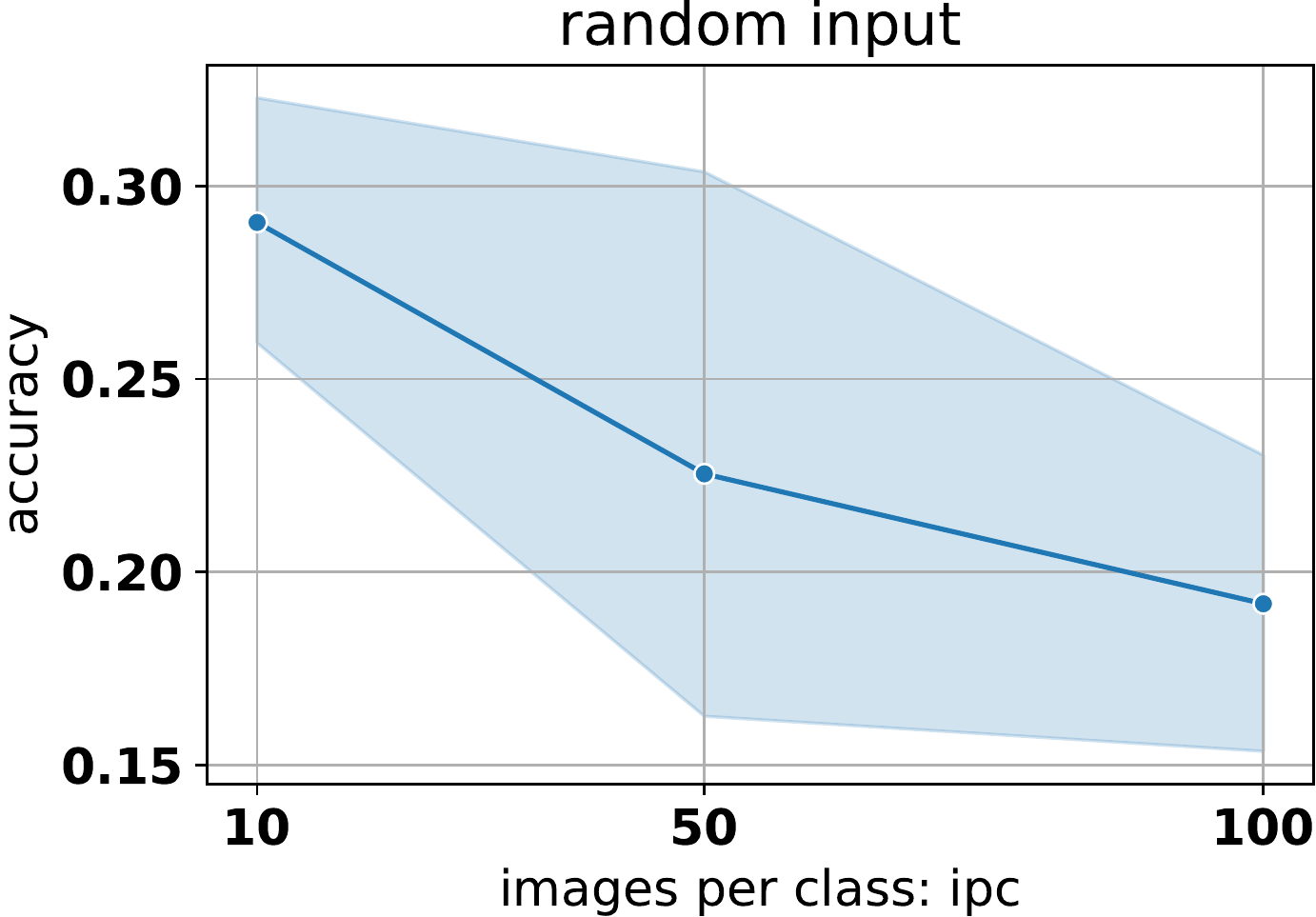}\\e)
\end{minipage}
\hfill
\begin{minipage}[h]{0.48\linewidth}
\centering
\includegraphics[width=\linewidth]{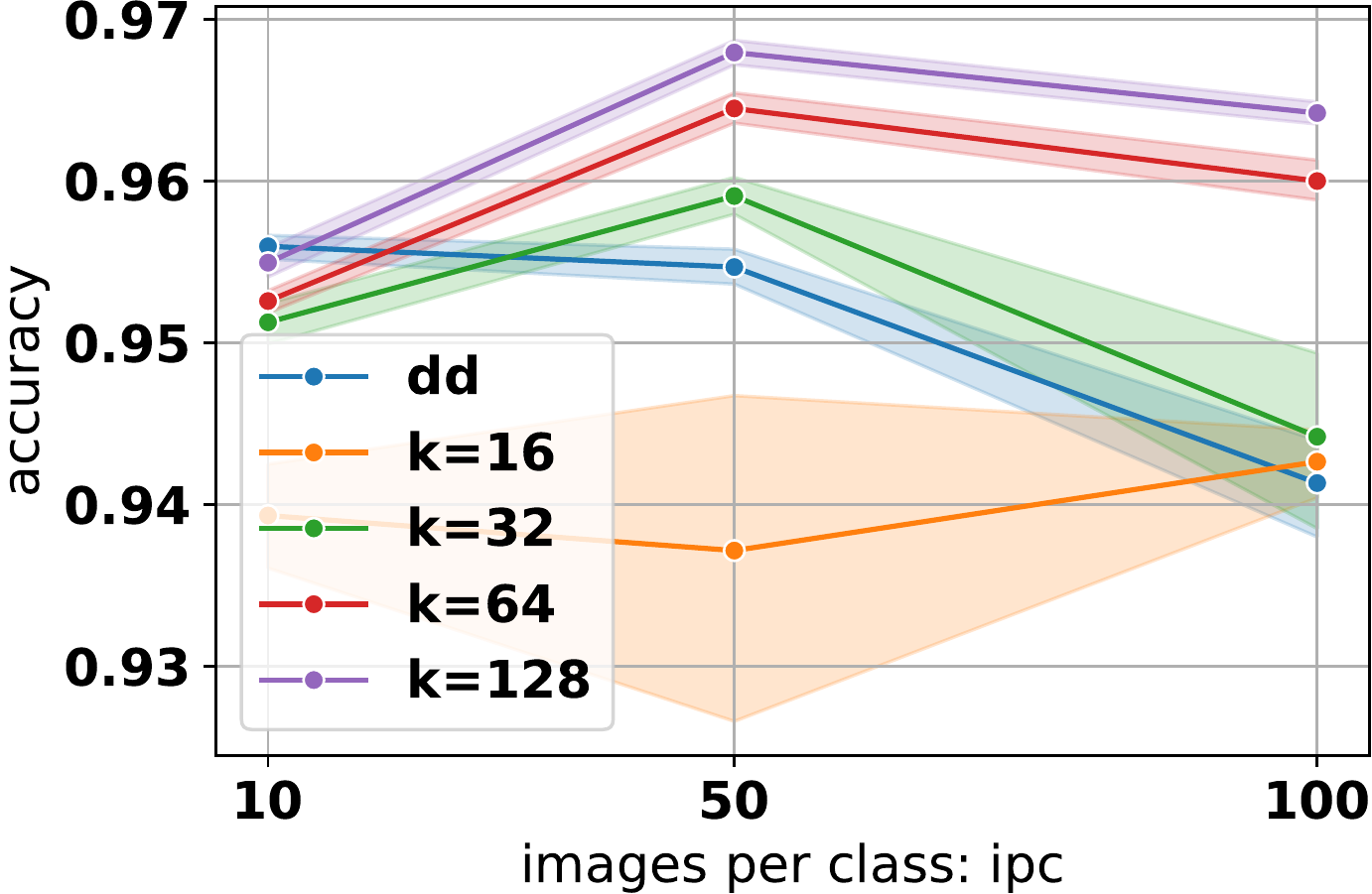}\\f)
\end{minipage}
\caption{Dependence of student's performance and hyperparameters of distillation procedure. Next parameters used as default: $ipc=10, ic=1, N=10, \zeta_\theta=10, k=64$.}\label{ris1}
\end{figure}

In this section we explore the use of gradient matching to train teacher generator. We first check the hyperparameters for this distillation method. $N$ controls frequency of student's reinitialization, $\zeta_{\theta}$ controls the speed at which teacher parameters are updated. \figurename~\ref{ris1} (a-d) shows the non-trivial relationship between performance and hyperparameter choice. We assume that such a dependence can be caused by the time limit and the fact that increasing the values of these hyperparameters may cause longer convergence. Note that in previous works~\cite{l1},~\cite{l15},~\cite{l16} where no time limit was used, increasing \textit{ipc} always resulted in better performance.
 
\figurename \ref{ris1}.e shows that fixation the generator input is really important for gradient matching distillation because teacher $\lambda$ training diverges when using random input. Another important detail mentioned above is that the gradient must be calculated per class. Table~\ref{tab2} shows the results for per class case and not. It seems that per class distillation gives significantly better results.

\figurename~\ref{ris1}.f shows the accuracy achieved with data distilled with generators of different sizes (marked with different $k$), and without a generator (\textbf{DD}). This plot depicts the dependency between the number of synthetic images per class (\textit{ipc}) and student's performance on test. It seems that the correct size selection for the generator allows to get better performance. More detailed results can be found in Tables~\ref{tab2} and~\ref{tab3}. For experiment in Table~\ref{tab2}, we use $ipc=10,~ic=1,~N=10,~K=110,~\zeta_\theta=10$ and $k=64$ for GTN as default hyperparameters values. For experiment in Table~\ref{tab3}, we use $k=64, ipc=50, K=35, N=10,~\zeta_\theta=10$.

\begin{table}
\caption{Mean and standard deviation of test accuracy for different distillation algorithms.}
 \centering
 \begin{tabular}{l|c|c|c}
Method + Teacher & Accuracy & Params & GPU (MiB) \\\hline
GM + DD & \boldmath{$95.6 \pm 0.1$} & $78.4$ K & $\approx 2390$\\\hline
GM + DD & $86.9 \pm 1.5$ & $78.4$ K & $\approx 2370$ \\
(not per class) &&&\\\hline
GM + GTN-lrn & \boldmath{$95.2 \pm 0.1$} & $1.646$ M & $\approx 2454$ \\\hline
GM + GTN-lrn & $93.4 \pm 0.3$ & $1.646$ M & $\approx 2434$\\
(not per class) &&&\\\hline
 \end{tabular}
 \label{tab2}
\end{table}

\begin{table}[h!]
\caption{Mean and standard deviation of test accuracy for different distillation algorithms.}\label{tab3}
 \centering
 \begin{tabular}{l|c|c|c}
Method + Teacher & Accuracy & Params & GPU (MiB) \\\hline
GM + GTN-lrn  & $94.2 \pm 0.4$ & $172.2$ K & \boldmath{$\approx 4192$} \\
($k=16, ipc=100$)&&&\\\hline
GM + GTN-lrn & $95.9 \pm 0.2$ & $449.7$ K & $\approx 3610$ \\
($k=32, K=50$)&&&\\\hline
GM + GTN-lrn & $96.4 \pm 0.1$ & $1.672$ M & $\approx 3640$ \\
($K=50$)&&&\\\hline
GM + GTN-lrn & \boldmath{ $96.8 \pm 0.1$} & \boldmath{$6.533$} M & $\approx 3770$ \\
$(k=128, K=50$)&&&\\\hline
GM + GTN-rnd & $29.0 \pm 6.1$ & $1.640$ M & $\approx 2454$\\
($ipc=10, K=110$)&&&\\\hline
 \end{tabular}
\end{table}

Tables~\ref{tab2} and~\ref{tab3} show the GPU memory usage. It seems that \textit{ipc} has a greater impact on memory usage than \textit{k}, which is another benefit of using \textbf{GTN}. Note that memory usage can be reduced by changing the \textit{ic} value to optimize more synthetic images using smaller batches. Note that such change can slow down convergence.
 
 \subsection{Distillation with implicit differentiation}\label{exp_ift}

\begin{figure}
\begin{minipage}[h]{0.48\linewidth}
\centering
\includegraphics[width=\linewidth]{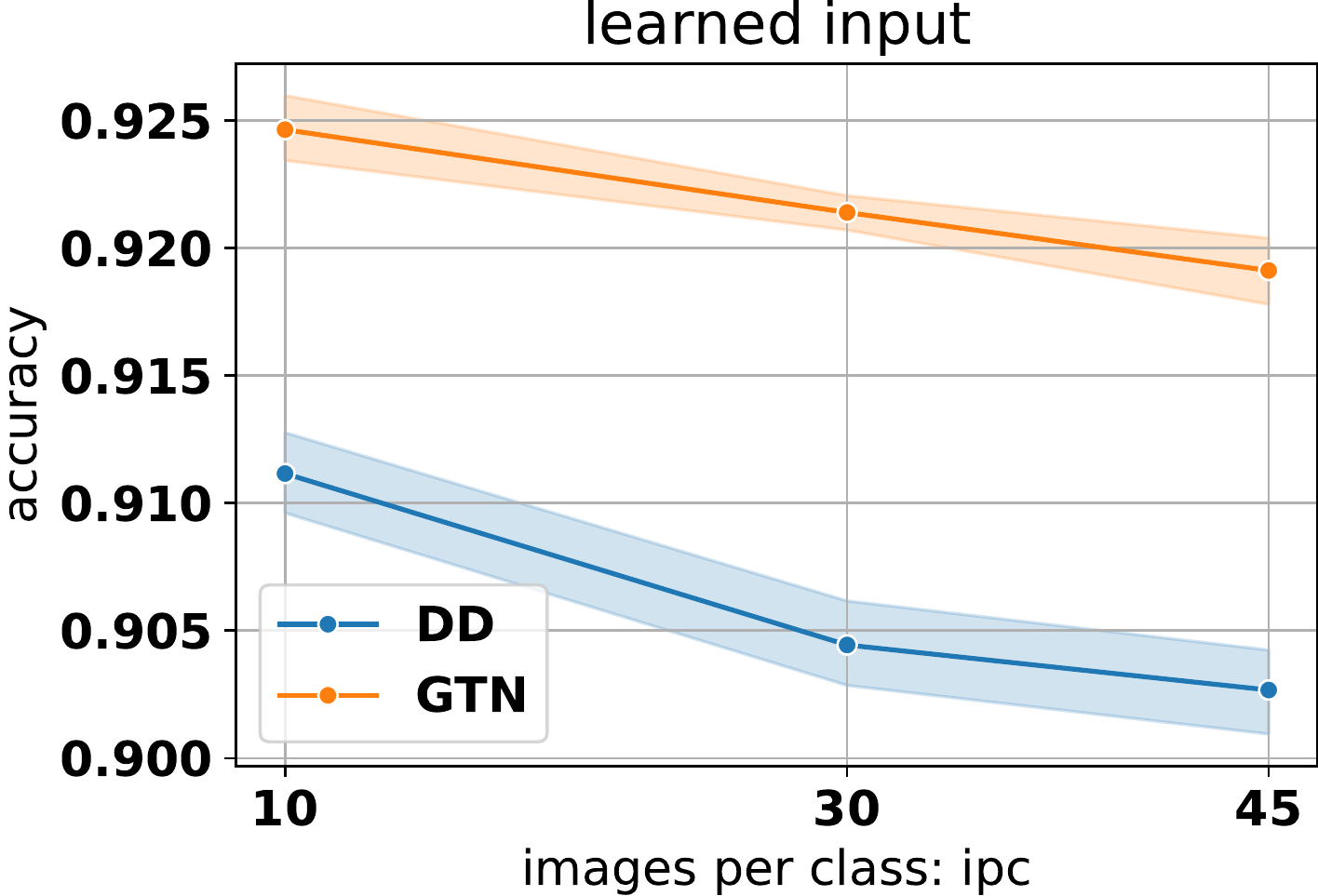}\\a)
\end{minipage}
\hfill
\begin{minipage}[h]{0.48\linewidth}
\centering
\includegraphics[width=\linewidth]{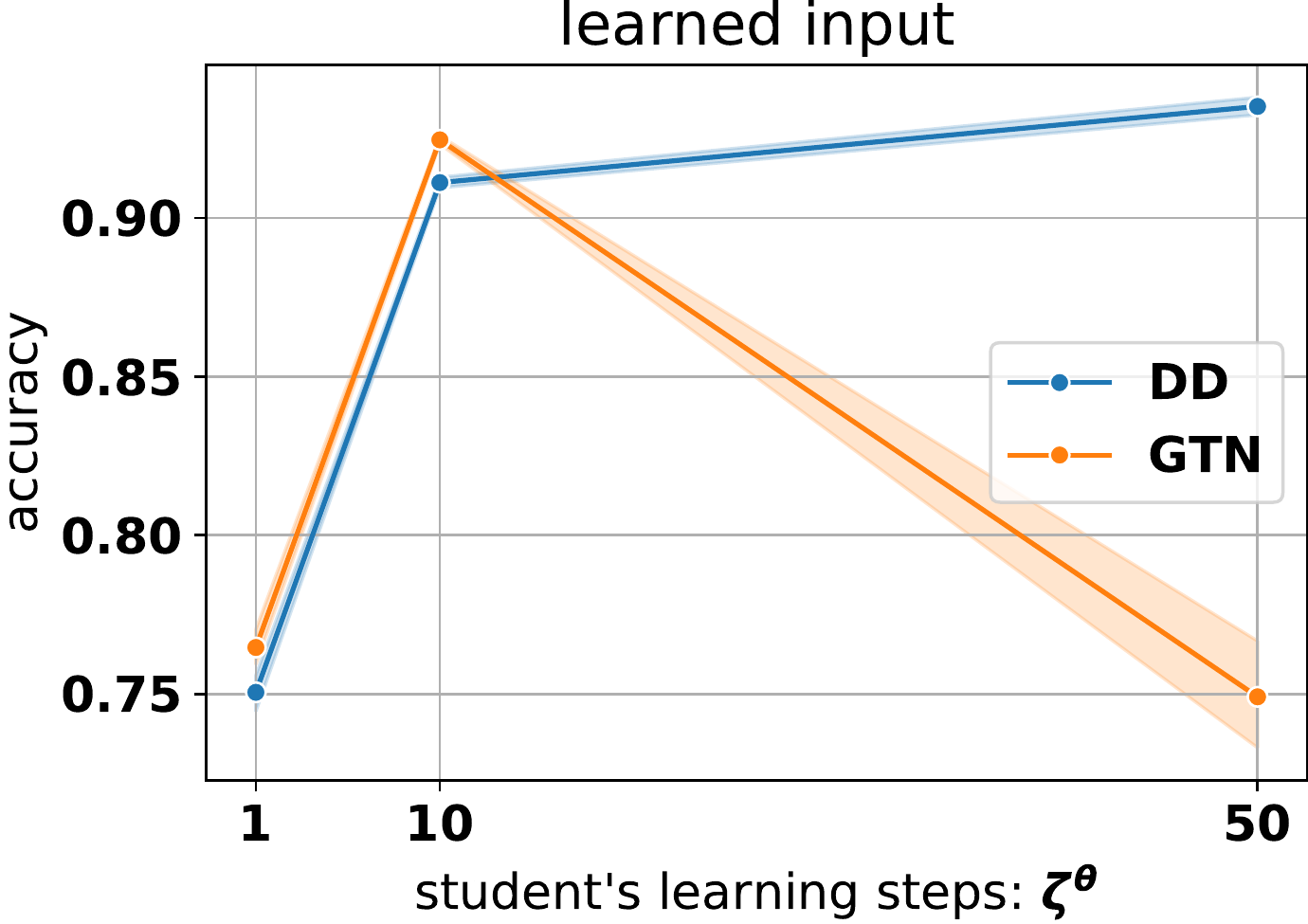}\\b)
\end{minipage}
\vfill
\begin{minipage}[h]{0.48\linewidth}
\centering
\includegraphics[width=\linewidth]{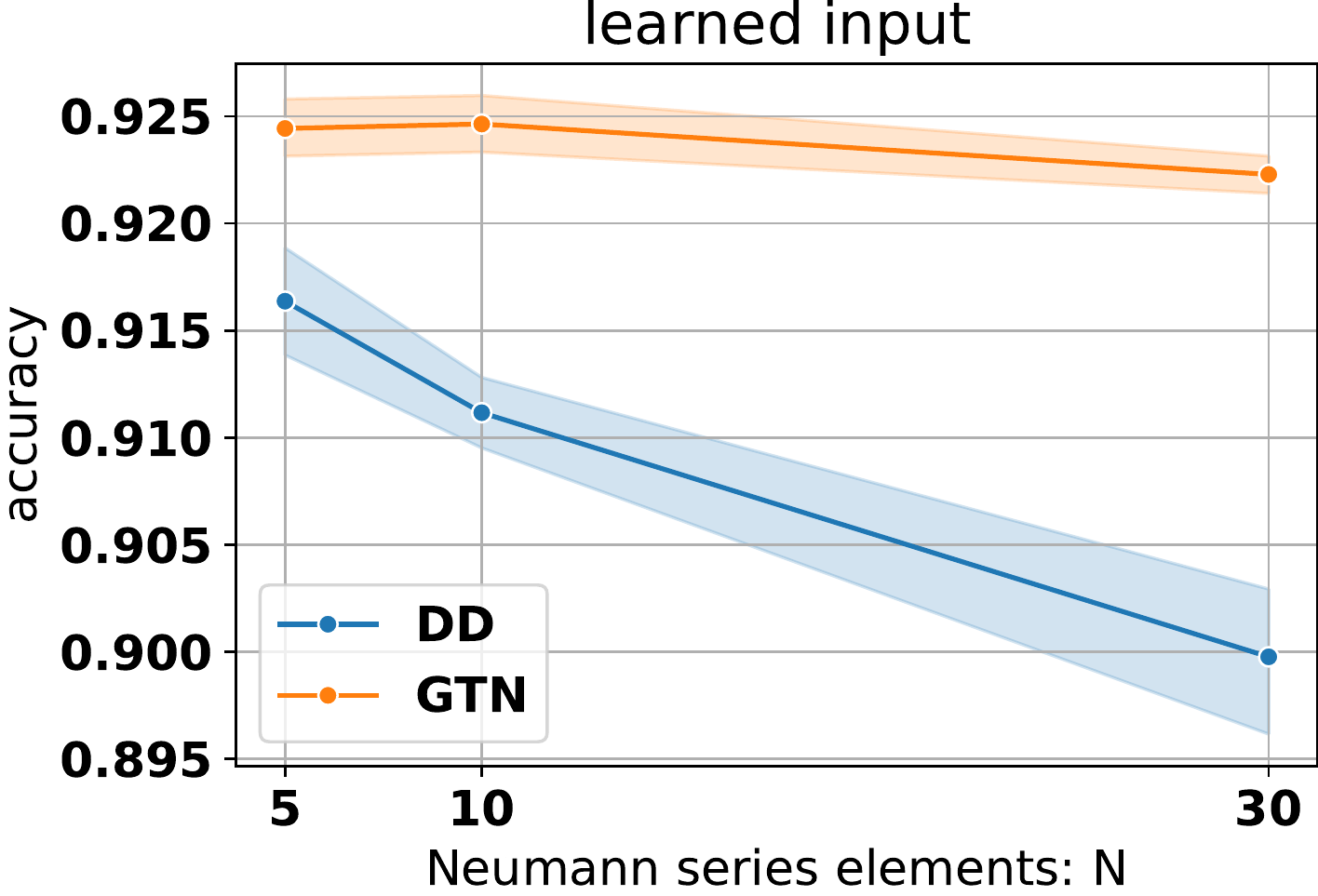}\\c)
\end{minipage}
\hfill
\begin{minipage}[h]{0.48\linewidth}
\centering
\includegraphics[width=\linewidth]{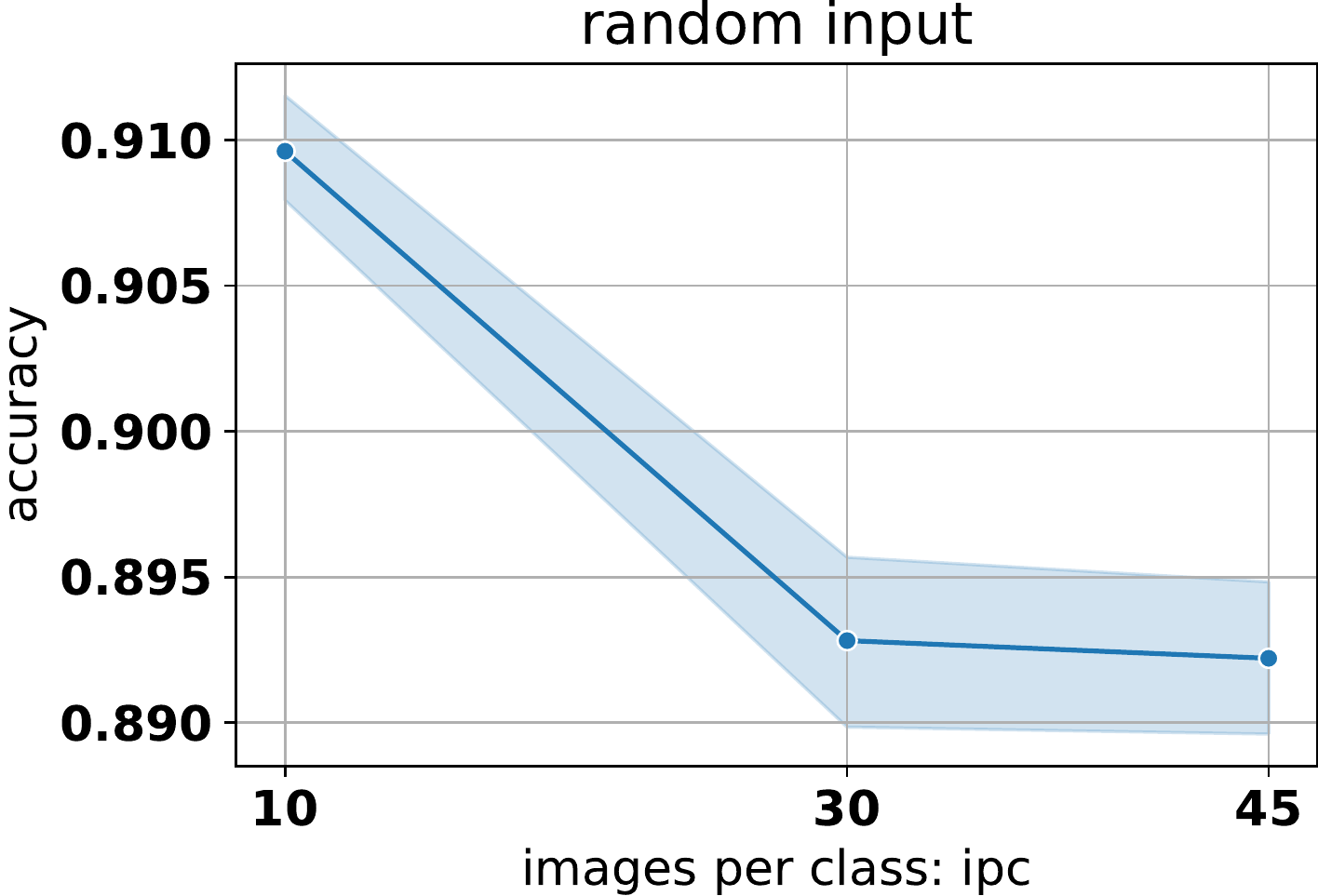}\\d)
\end{minipage}
\caption{relation of distillation method's hyperparameters and test performance. We use as default: $ipc=10, N=10, \zeta_\theta=10, k=64$.}\label{ris2}
\end{figure}

This method was proposed in~\cite{l14}, and we will abbreviate it as \textbf{IFT} (implicit function theorem). As mentioned above (see section~\ref{impl_diff}), there is no detailed description of the results in the original paper, so they can be found in this section. \figurename~\ref{ris2} (a-c) shows the relationship between the hyperparameters of the distillation method and the student's performance on the test. We assume that these results can be explained by the fact that increasing the values of these hyperparameters decreases frequency of $\lambda$ update, which negatively affects performance. The only exception is $\zeta_{\theta}$. 

\figurename~\ref{ris2}.d shows results for distillation using generator with random input (\textbf{GTN-rnd}). Such a generator can produce as much data as we need, but it can't converge when trained with gradient matching. It seems that such distillation becomes possible using implicit differentiation.

Table~\ref{tab4} shows the best results for each method. For this experiment, we use $K=1080, \zeta_\theta=50, ipc=10,~N=10$ as default hyperparameters values. The performance seems to be the same or even better compared to backpropagation through the training procedure \textbf{unroll} (see. Table~\ref{tab1}). Note the difference in memory usage in both tables. Note that the implicit differentiation distillation is inferior to the gradient matching distillation.
We think this may be connected with the difference in frequency of $\lambda$ update. To do one update using \textbf{IFT}, we first have to train the student, which is not needed in case of \textbf{GM}. It is also important to note that this method is very sensitive to $\alpha$ and $\zeta_{\theta}$, and in some \textbf{DD} cases it starts to diverge after several iterations. Meanwhile the use of \textbf{GTN} makes the procedure more stable and allows for a more generalizable dataset (see Table~\ref{tab6}).

\begin{figure}[h!]
\begin{minipage}[h]{.49\linewidth}
\centering
\includegraphics[width=\linewidth]{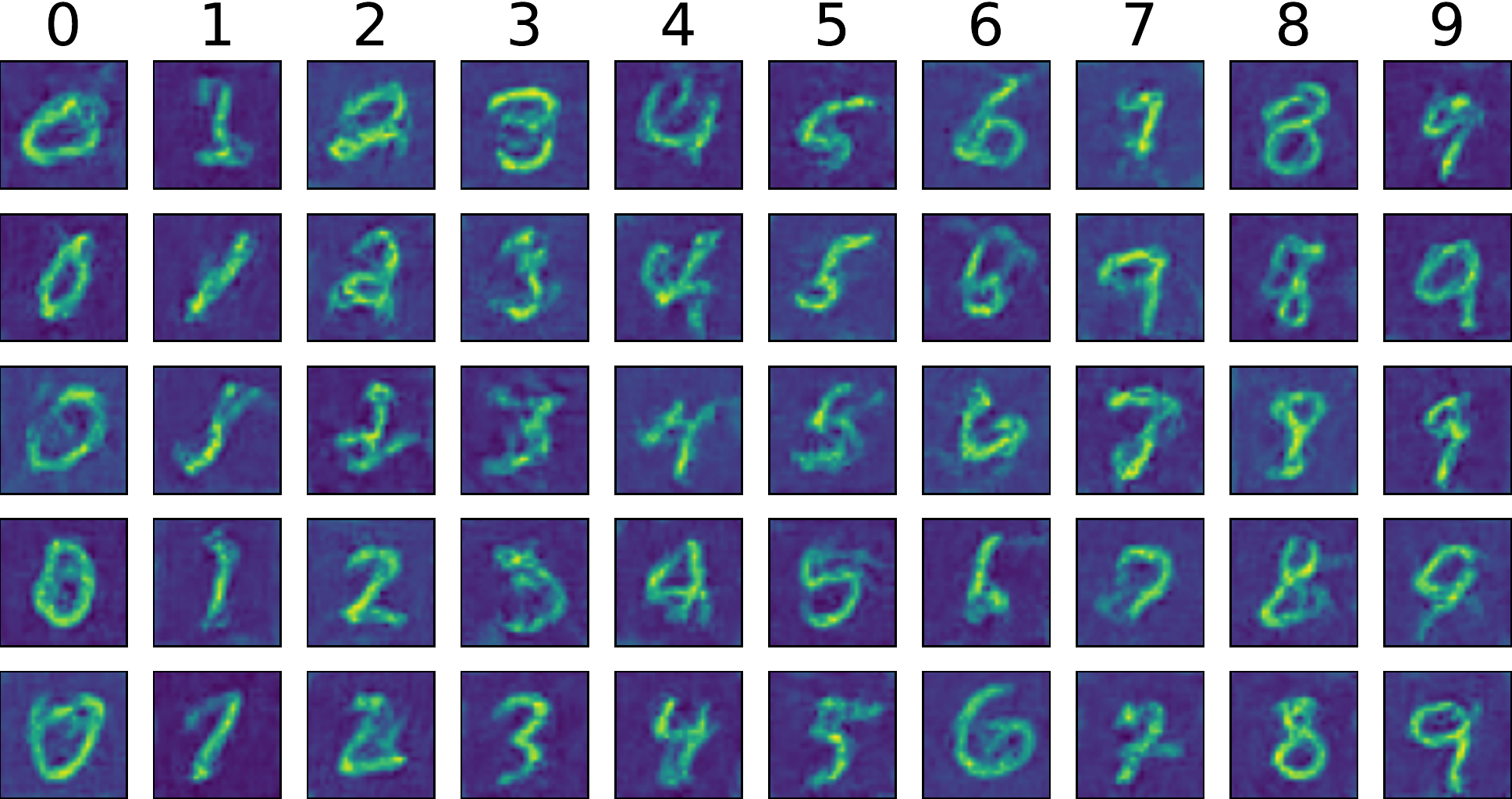}\\a)
\end{minipage}
\hfill
\begin{minipage}[h]{.49\linewidth}
\centering
\includegraphics[width=\linewidth]{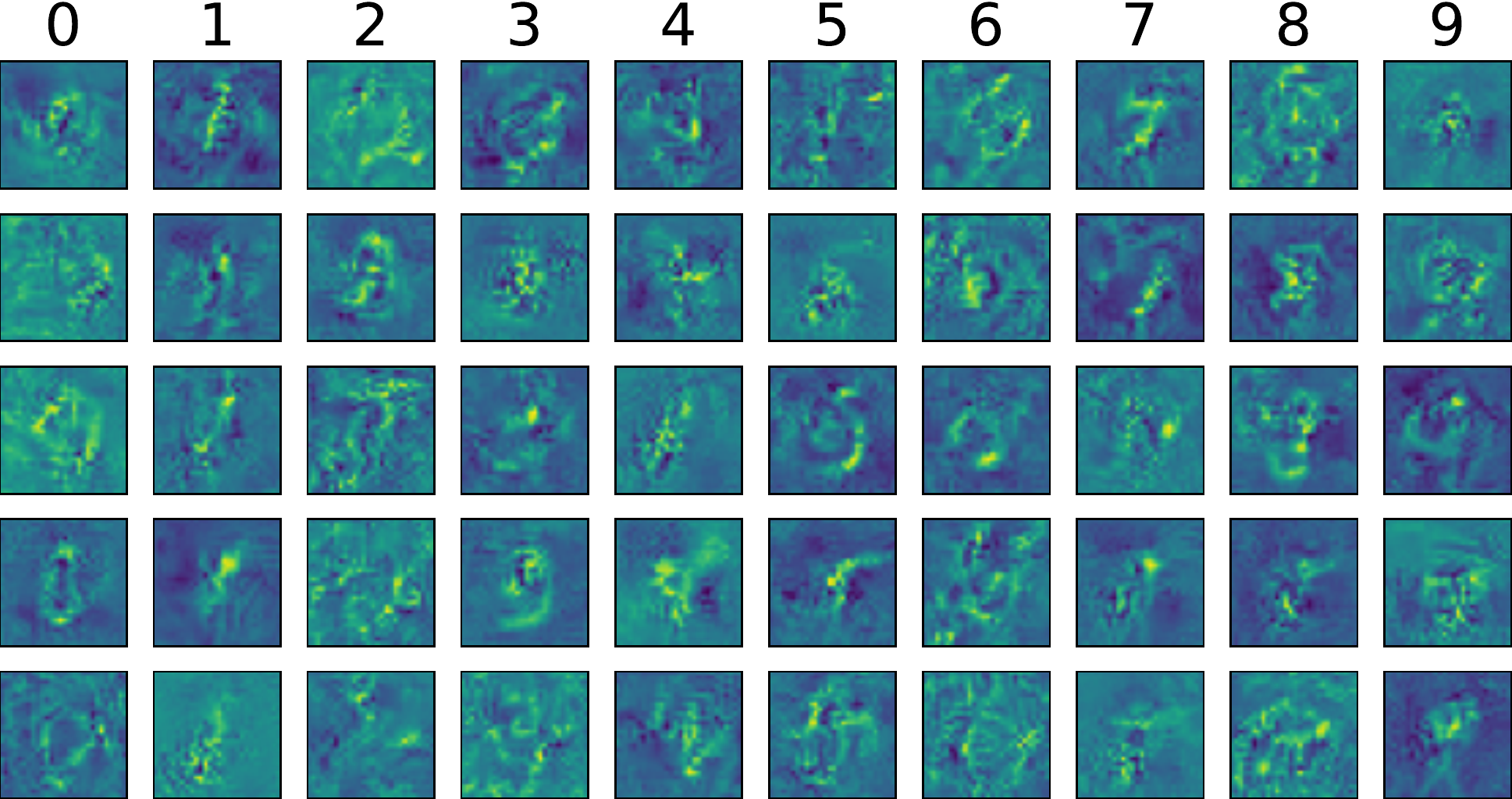}\\b)
\end{minipage}
\vfill
\begin{minipage}[h]{.49\linewidth}
\centering
\includegraphics[width=\linewidth]{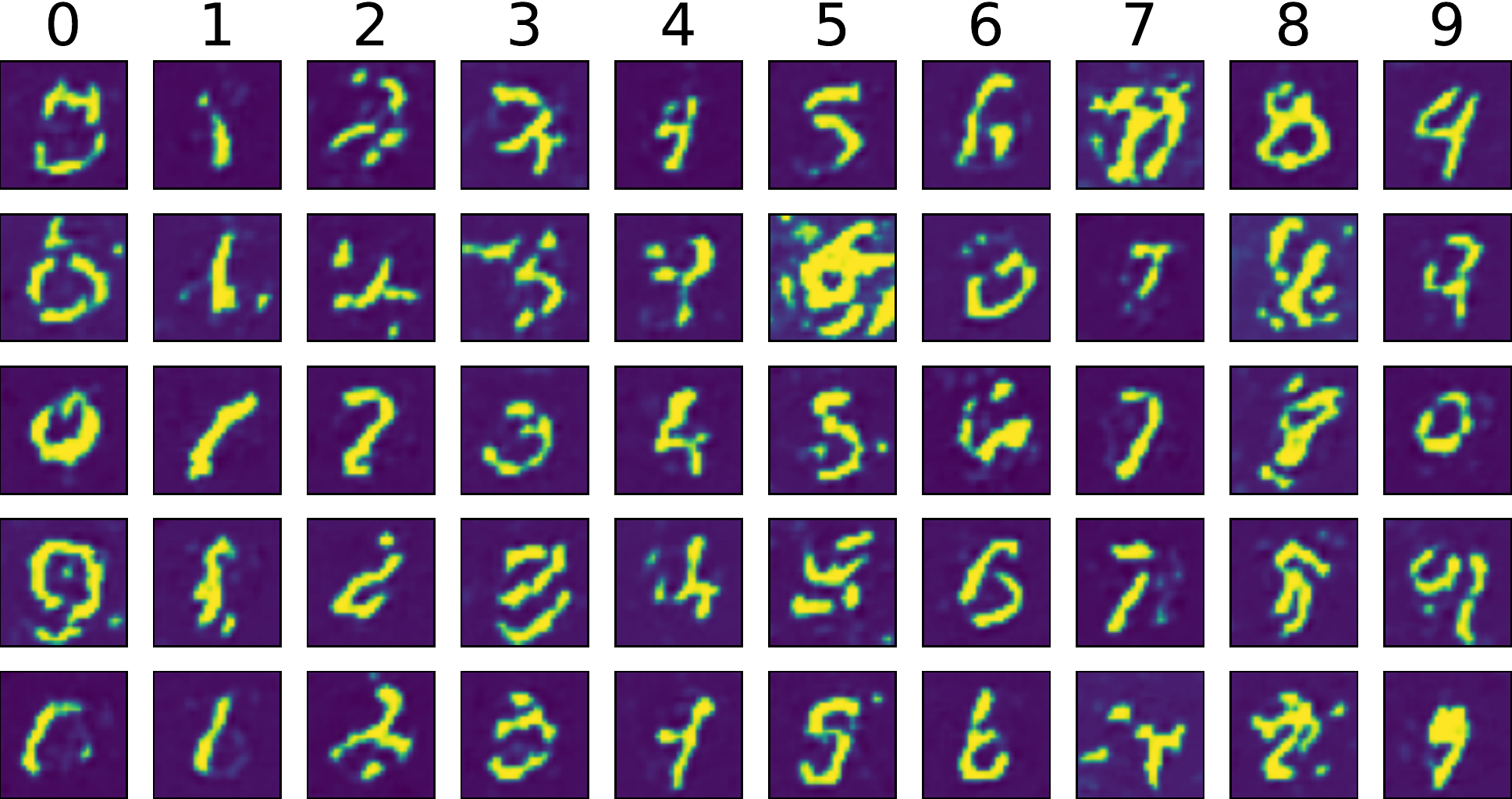}\\c)
\end{minipage}
\hfill
\begin{minipage}[h]{.49\linewidth}
\centering
\includegraphics[width=\linewidth]{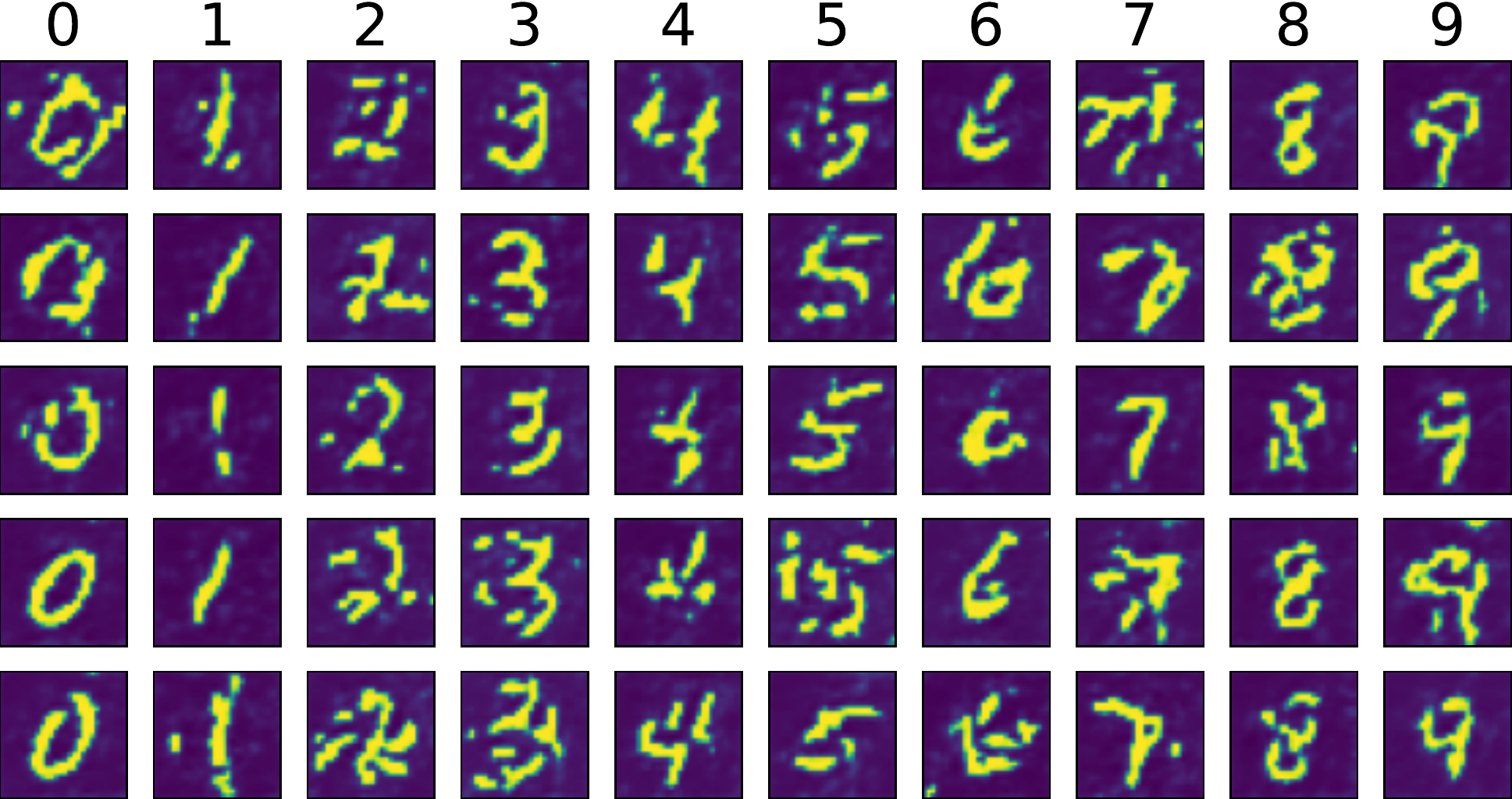}\\d)
\end{minipage}
\vfill
\begin{minipage}[h]{.49\linewidth}
\centering
\includegraphics[width=\linewidth]{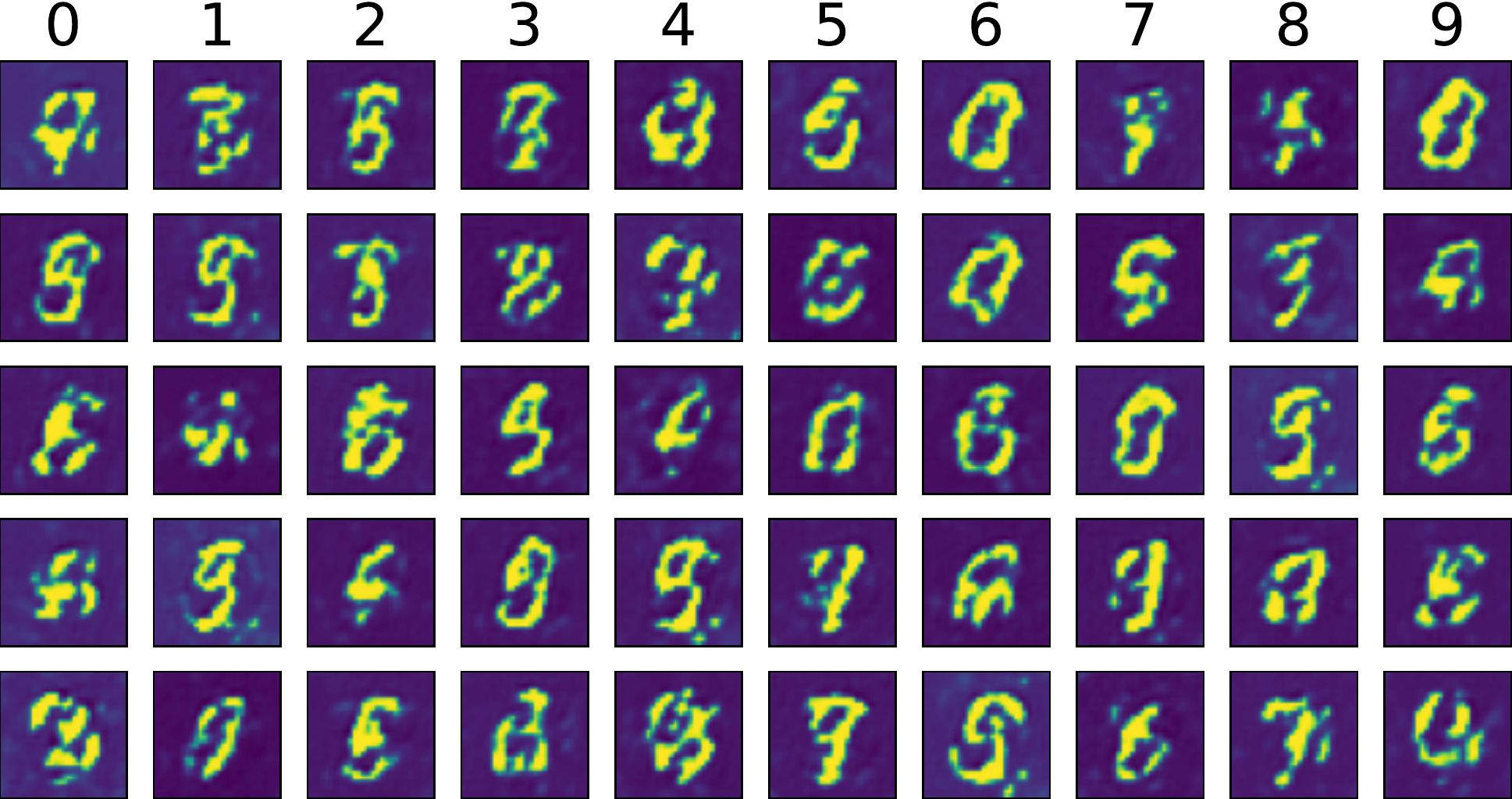}\\e)
\end{minipage}
\hfill
\begin{minipage}[h]{.49\linewidth}
\centering
\includegraphics[width=\linewidth]{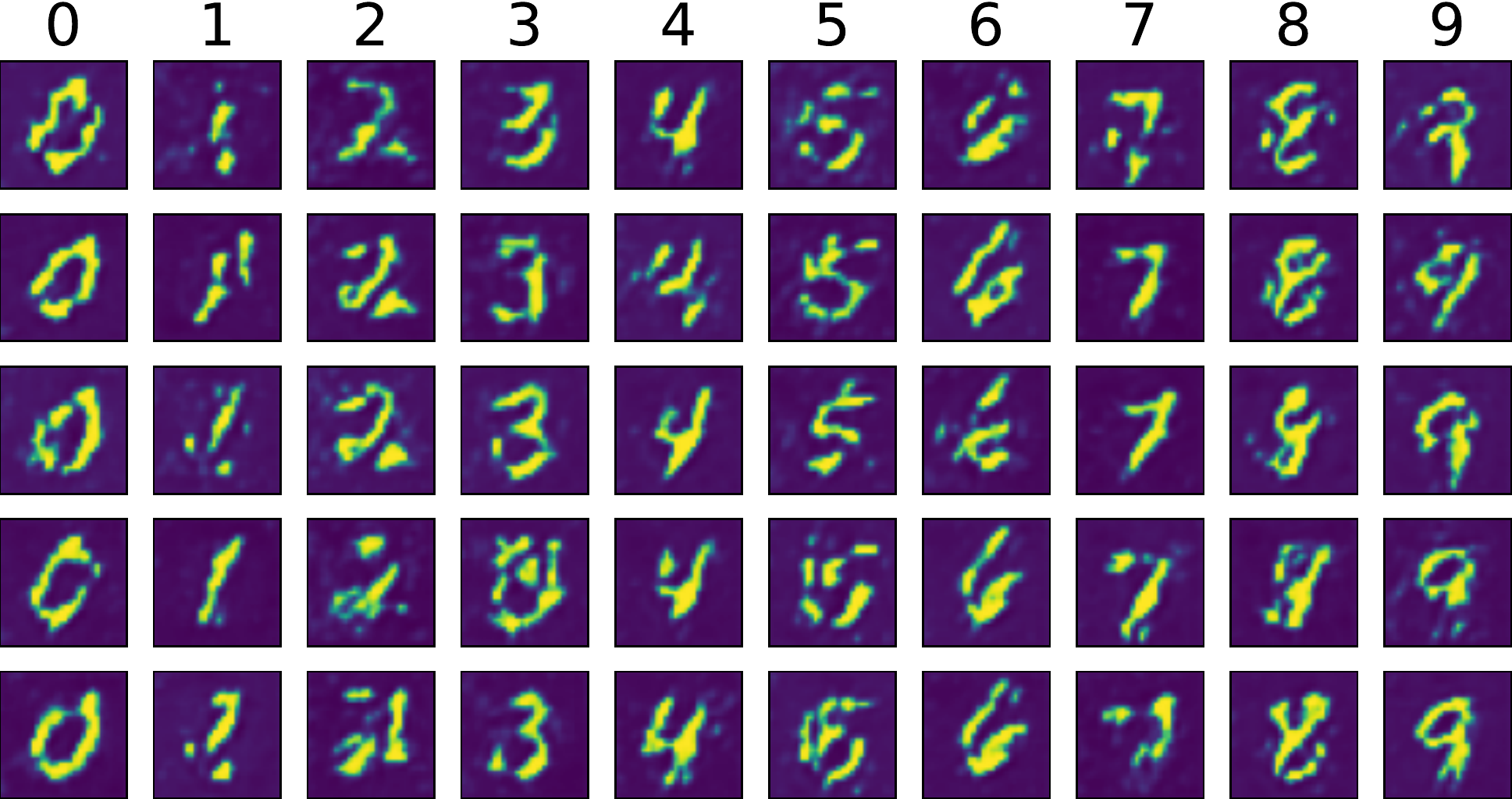}\\f)
\end{minipage}
\caption{Synthetic images for MNIST classification task obtained with different distillation methods: a) GM+DD, b) IFT+DD, c) GM+GTN-lrn, d) IFT+GTN-lrn, e) GM+GTN-rnd, f) IFT+GTN-rnd. We use the same hyperparameters as mentioned in table~\ref{tab5}. Hyperparameters for GM+GTN-rnd are described in caption of table~\ref{tab4}.}\label{ris3}
\end{figure}

\begin{table}[h!]
\caption{Mean and standard deviation of test accuracy for different distillation algorithms.}
 \centering
 \begin{tabular}{l|c|c|c}
Method + Teacher & Accuracy & Params & GPU (MiB) \\\hline
IFT + DD ($K=500$) & \boldmath{$93.5 \pm 0.5$} & $78.4$ K & $\approx 2726$\\\hline
IFT + GTN-lrn ($\zeta_\theta=10$) & $92.4 \pm 0.2$ & \boldmath{$1.646$} M & $\approx 2726$\\\hline
IFT + GTN-rnd ($\zeta_\theta=10$) & $90.9 \pm 0.3$ & $1.640$ M & $\approx 2726$\\\hline
 \end{tabular}
\label{tab4}
\end{table}

\figurename~\ref{ris3} shows part of the final synthetic dataset for \textbf{GM} (see a, c, e) and \textbf{IFT} (see b, d, f). The greatest difference is obtained when data distilled without a generator (see a, b). Synthetic data obtained using implicit differentiation looks less realistic and therefore can be used for federative learning~\cite{l20}. Also note that the images distilled using generator have more contrast.

\subsection{Distillation with augmentation}\label{exp_aug}

In previous works, augmentation has been used in different ways. In~\cite{l16} it takes place during distillation (let's call it train augmentation) by applying transformations to real images $\mathcal{B^T}$. In~\cite{l1},~\cite{l15} it is used when teaching student on synthetic data (let's call it test augmentation). In our study, we decided to compare augmentation techniques. Table~\ref{tab5} shows the test performance for various distillation and augmentation techniques. It seems that for the MNIST classification problem only test augmentation gives improvement (see tables~\ref{tab2},~\ref{tab3},~\ref{tab4}). To augment images we use random crop and rotation. For this experiment, we use $K=1080, ipc=10,~\zeta_\theta=10,~N=10$ as default hyperparameters values.

\begin{table}[h!]
\caption{Mean and standard deviation of test accuracy for different distillation algorithms and different augmentations.}
 \centering
 \begin{tabular}{l|c|c|c}
Method +& Test Aug. & Train Aug. & Test + \\
Teacher & & & Train Aug. \\\hline
GM+DD ($ic=1,$& $96.1 \pm 0.4$ & $94.8 \pm 0.1$ & $93.9 \pm 0.5$ \\
$\hspace{1cm}K=110$)&&&\\\hline
GM+GTN-lrn & \boldmath{$97.4 \pm 0.1$} & \boldmath{$96.2 \pm 0.2$} & \boldmath{$95.5 \pm 0.4$} \\
($k=128,ipc=50,$&&&\\
$K=50$)&&&\\\hline
IFT+DD ($\zeta_\theta=50,$& $92.3 \pm 0.9$ & $91.4 \pm 0.5$ & $89.2 \pm 1.5$ \\
$\hspace{1cm}K=500$)&&&\\\hline
IFT+GTN-lrn & $93.0 \pm 0.2$ & $91.4 \pm 0.3$ & $91.4 \pm 0.4$ \\\hline
IFT+GTN-rnd & $92.2 \pm 0.3$ & $89.7 \pm 0.3$ & $90.9 \pm 0.6$ \\\hline
 \end{tabular}
\label{tab5}
\end{table}

\subsection{Generalizability}\label{exp_gen}

The generalization problem of distilled data was first mentioned in~\cite{l1} and then studied in~\cite{l16} and~\cite{l17}. The problem is that such data can't guarantee convergence for students which didn't participate in the distillation procedure. And this problem is of great importance, since the main practical use of synthetic data is NAS. For this experiment, we use $K=1080, ipc=10,~\zeta_\theta=10,~N=10$ as default hyperparameters values.

\begin{table}
\caption{Mean and standard deviation of test accuracy for different distillation algorithms and student's architectures.}
 \centering
 \begin{tabular}{l|c|c|c|c}
Method + & LeNet & AlexNet & VGG11 & MLP\\
Teacher &&&&\\\hline
GM+DD & $94.1 \pm 0.6$ & $95.0 \pm 0.2$ & $95.8 \pm 0.3$ & \boldmath{$88.6$} \\
&&&&$\pm 0.4$\\\hline 
GM+ & \boldmath{$95.5 \pm 0.3$} & \boldmath{$96.7 \pm 0.2$} & \boldmath{$97.4 \pm 0.1$} & $86.8$ \\
GTN-lrn &&&&$\pm 0.3$\\\hline
IFT+DD & $74.0 \pm 7.8$ & $68.6 \pm 8.9$ & $86.5 \pm 1.6$ & $50.9$ \\
&&&&$\pm 8.3$\\\hline 
IFT+ & $91.5 \pm 1.0$ & $82.5 \pm 14.9$& $93.0 \pm 0.4$& $79.9$ \\
GTN-lrn &&&&$\pm 0.6$\\\hline
IFT+ & $88.3 \pm 2.3$ & $85.3 \pm 3.9$ & $92.1 \pm 0.4$ & $74.4$ \\
GTN-rnd &&&&$\pm 1.1$\\\hline
 \end{tabular}
\label{tab6}
\end{table}

Table~\ref{tab6} shows the results of students with different architectures trained on data distilled with different methods. For distillation we used ConvNet student's architecture, all results were obtained with test augmentation. It seems that the best generalizability can be obtained using \textbf{GTN} and \textbf{GM} use. For a comparison with ConvNet see the first column of Table~\ref{tab5}. 

\section{Conclusion}\label{conc}
This work explores all the latest ideas in dataset distillation field suggested in~\cite{l1},~\cite{l14},~\cite{l15},~\cite{l16}. We honestly compared the performance of all known methods, limiting their running time. We also proposed new methods based on the joint use of generators and memory efficient methods. Experiments with the MNIST benchmark show that selecting the correct size for the generator allows to achieve better performance for gradient matching distillation, and improves the generalizability of implicit differentiation distillation. This paper also presents the results of augmentation impact on distillation. We also provide a detailed description of the experimental results for implicit differentiation distillation, as we didn't find them in the original work~\cite{l14}. As future work, we want to experiment with much more diverse datasets and architectures. We also want to improve the distilled data generalizing ability using stochastic depth networks~\cite{l13}. We are also interested in experiments with bringing the distribution of synthetic objects closer to the original one.

\section*{Acknowledgment}
\addcontentsline{toc}{section}{Acknowledgment}
This research was performed at the Center for Big Data Storage and Analysis of Lomonosov Moscow State University and was supported by the National Technology Initiative Foundation (13/1251/2018 of December 11, 2018).


\begin{thebibliography}{00}
\bibitem{l1} Wang\,T., Zhu\,J., Torralba\,A., Efros\,A. A.: Dataset Distillation. CoRR; abs/1811.10959 (2018)

\bibitem{l14} Lorraine\,J., Vicol\,P., Duvenaud\,D. Optimizing Millions of Hyperparameters by Implicit Differentiation. CoRR; abs/1911.02590 (2019)

\bibitem{l15} Zhao\,B., Mopuri\,K.\,R., Bilen\,H. Dataset Condensation with Gradient Matching. CoRR; abs/2006.05929 (2020)

\bibitem{l16} Such\,F.\,P., Rawal\,A., Lehman\,J., Stanley\,K.\,O., Clune\,J. 
Generative Teaching Networks: Accelerating Neural Architecture Search by Learning to Generate Synthetic Training Data. CoRR; abs/1912.07768 (2019)

\bibitem{l8} Maclaurin\,D., Duvenaud\,D. and Adams\,R.: Gradient-Based Hyperparameter Optimization Through Reversible Learning. CoRR; abs/1502.03492 (2015)

\bibitem{l3} Sucholutsky\,I., Schonlau\,M.: Soft-Label Dataset Distillation and Text Dataset Distillation. CoRR; abs/1910.02551 (2019)

\bibitem{l17} Medvedev\,D., D’yakonov\,A. New Properties of the Data Distillation Method When Working with Tabular Data. In: van der Aalst W.M.P. et al. (eds) Analysis of Images, Social Networks and Texts. AIST 2020. Lecture Notes in Computer Science, vol 12602. Springer, Cham. CoRR; abs/2010.09839 (2021)

\bibitem{l6} LeCun\,Y., Boser\,B., Denker\,J. S., Henderson\,D., Howard\,R. E., Hubbard\,W., and Jackel\,L. D.: Backpropagation Applied to Handwritten Zip Code RecognitionNeural Computation. Neural Computation 1(4), 541--551 (1989)

\bibitem{l9} Bengio\,Y.: Gradient-Based Optimization of Hyperparameters. Neural Computation 12(8), 1889--1900 (2000)

\bibitem{l10} Baydin\,A., Pearlmutter\,B.: Automatic Differentiation of Algorithms for Machine Learning. In: Proceedings of the AutoML Workshop at the International Conference on Machine Learning (ICML). Beijing, China, June 21--26 (2014) 

\bibitem{l11} Liu\,D. C., Nocedal\,J.: On the Limited Memory BFGS Method for Large Scale Optimization. Mathematical Programming 45, 503--528 (1989)

\bibitem{l12} Polyak\,B.: Some Methods of Speeding Up the Convergence of Iteration Methods. USSR Computational Mathematics and Mathematical Physics, vol. 4, pp. 1--17 (1964)

\bibitem{l7} Domke\,J.: Generic Methods for Optimization-Based Modeling. In: Proceedings of the Fifteenth International Conference on Artificial Intelligence and Statistics, pp. 318--326. PMLR (2012)

\bibitem{l4} MNIST Handwritten Digit Database, \url{http://yann.lecun.com/exdb/mnist/}. Last accessed 17 April 2021.

\bibitem{l5} Lecun\,Y., Bottou\,L., Bengio\,Y., Haffner\,P.: Gradient-Based Learning Applied to Document Recognition. In: Proceedings of the IEEE, vol. 86, pp. 2278--2324 (1998)

\bibitem{l2} Hinton\,G., Vinyals\,O., Dean\,J.: Distilling the Knowledge in a Neural Network. In: NIPS Deep Learning and Representation Learning Workshop. (2015)

\bibitem{l18} Grefenstette\,E., Amos\,B., Yarats\,D., Htut\,P.\,M., Molchanov\,A., Meier\,F., Kiela\,D., Cho\,K., Chintala\,S. Generalized Inner Loop Meta-Learning. CoRR; abs/1910.01727 (2019)

\bibitem{l13} Huang\,G., Sun\,Y., Liu\,Z., Sedra\,D. and Weinberger\,K.: Deep Networks With Stochastic Depth. CoRR; abs/1603.09382 (2016)

\bibitem{l19} Gidaris\,S., Komodakis\,N. Dynamic Few-Shot Visual Learning without Forgetting. CoRR; abs/1804.09458 (2018)

\bibitem{l20} Zhou\,Y., Pu\,G., Ma\,X., Li\,X., Wu\,D. Distilled One-Shot Federated Learning. CoRR; abs/2009.07999 (2020)

\end{thebibliography}
\end{document}